\documentclass[a4paper]{article}
\usepackage[a4paper, total={6.5in, 8.66in}]{geometry}

\usepackage{authblk}
\usepackage[ruled,vlined]{algorithm2e}
\SetKwComment{Comment}{$\triangleright$\ }{}   
\SetAlgoHangIndent{1.2em}          

\usepackage{amsmath}
\usepackage{upgreek}

\usepackage{subcaption}

\usepackage[utf8]{inputenc}
\usepackage{amsmath}
\usepackage{amssymb}
\usepackage{mathtools}
\usepackage{mathrsfs}

\usepackage{siunitx}
\usepackage[colorinlistoftodos]{todonotes}

\usepackage[toc,page]{appendix}

\usepackage[hidelinks]{hyperref}

\usepackage{booktabs}  
\usepackage{multirow}  
\newcommand{\papertitle}{Shape-informed cardiac mechanics surrogates in data-scarce regimes via geometric encoding and generative augmentation}

\DeclareMathOperator*{\argmin}{arg\,min}

\newcommand{\avsum}{\mathop{\mathpalette\avsuminner\relax}\displaylimits}

\makeatletter
\newcommand\avsuminner[2]{%
  {\sbox0{$\m@th#1\sum$}%
   \vphantom{\usebox0}%
   \ooalign{%
     \hidewidth
     \smash{\vrule height\dimexpr\ht0+1pt\relax depth\dimexpr\dp0+1pt\relax}%
     \hidewidth\cr
     $\m@th#1\sum$\cr
   }%
  }%
}
\makeatother

\newcommand{\shapecode}{\mathbf{z}}
\newcommand{\geoparam}{\boldsymbol{\mu}}
\newcommand{\point}{\mathbf{x}}

\newcommand{\CD}{\text{CD}}
\newcommand{\CDnorm}{\text{CD}_{\text{norm}}}
\newcommand{\IoU}{\text{IoU}}
\newcommand{\Dice}{\text{Dice}}
\newcommand{\MLE}{\text{MLE}}

\newcommand{\SDFmodel}{\text{SDF-SM}}
\newcommand{\PCAmodel}{\text{PCA-SM}}

\newcommand{\SDFusm}{\text{SDF-USMNet}}
\newcommand{\PCAusm}{\text{PCA-USMNet}}

\newcommand{\SDF}{\text{SDF}}

\newcommand{\NNgeneric}{\mathcal{N\!N}}
\newcommand{\NNsdf}{\NNgeneric_{\mathrm{sdf}}}
\newcommand{\weightNN}{\mathbf{w}}
\newcommand{\NNphysics}{\NNgeneric_{\mathrm{phys}}}
\newcommand{\weightNNsdf}{\weightNN_{\mathrm{sdf}}}
\newcommand{\weightNNphysics}{\weightNN_{\mathrm{phys}}}

\newcommand{\e}[1]{\times 10^{#1}}

\graphicspath{{img/}}

\title{{\papertitle}}
\date{}

\begin{document}
    \author[1]{Davide Carrara\textsuperscript{*}}
    \author[1,2]{Marc Hirschvogel\textsuperscript{*}}
    \author[1]{Francesca Bonizzoni}
    \author[1]{Stefano Pagani}
    \author[3]{Simone Pezzuto}
    \author[1]{Francesco Regazzoni}
    \affil[1]{\footnotesize{MOX Laboratory, Department of Mathematics, Politecnico di Milano, Milano, Italy}}
    \affil[2]{\footnotesize{Division 2.2 Process Simulation, Bundesanstalt für Materialforschung und -prüfung (BAM), Berlin, Germany}}
    \affil[3]{\footnotesize{Department of Mathematics, University of Trento, Trento, Italy}}
	\maketitle

    \begingroup
    \renewcommand\thefootnote{*}
    \footnotetext{These authors equally contributed to this work.}
    \endgroup

	\begin{abstract}
		High-fidelity computational models of cardiac mechanics provide mechanistic insight into the heart function but are computationally prohibitive for routine clinical use. Surrogate models can accelerate simulations, but generalization across diverse anatomies is challenging, particularly in data-scarce settings. We propose a two-step framework that decouples geometric representation from learning the physics response, to enable shape-informed surrogate modeling under data-scarce conditions. First, a shape model learns a compact latent representation of left ventricular geometries. The learned latent space effectively encodes anatomies and enables synthetic geometries generation for data augmentation. Second, a neural field-based surrogate model, conditioned on this geometric encoding, is trained to predict ventricular displacement under external loading. The proposed architecture performs positional encoding by using universal ventricular coordinates, which improves generalization across diverse anatomies. Geometric variability is encoded using two alternative strategies, which are systematically compared: a PCA-based approach suitable for working with point cloud representations of geometries, and a DeepSDF-based implicit neural representation learned directly from point clouds. Overall, our results, obtained on idealized and patient-specific datasets, show that the proposed approaches allow for accurate predictions and generalization to unseen geometries, and robustness to noisy or sparsely sampled inputs.
	\end{abstract}


    \section{Introduction}
    \label{sec:intro}

Computational models of the heart and the cardiovascular system have emerged as powerful tools to mechanistically investigate electrochemical, mechanical, and biological processes across multiple spatial and temporal scales \cite{quarteroni2015,schwarz2023}. However, high-fidelity models based on partial differential equations (PDEs) typically require computationally intensive discretization and solution techniques, often representing an obstacle to routine clinical translation and application.

To partially circumvent the computational burden without compromising accuracy, model order reduction techniques are often integrated into simulation frameworks. These reduce the problem to a lower-dimensional representation, either by exploiting snapshot information from high-fidelity models or imaging data \cite{hirschvogel2024-frsi,cicci2022,pfaller2020,farhat2014,boulakia2012}, or through lumped model approaches \cite{regazzoni2022b,quarteroni2016}.
However, real-time simulation through reduced-order models (ROMs) often remains computationally prohibitive, particularly in nonlinear regimes characterized by slow decay Kolmogorov $n$-width and a corresponding high-number of basis functions required to accurately represent the solution \cite{greif2019decay, pagani2018numerical}. 
This has motivated the development of data-driven surrogate models, which leverage dimensionality reduction techniques better suited for nonlinear solution manifolds \cite{lee2020model}. 
The resulting surrogate can be evaluated typically at a fraction of the computational cost compared to projection-based ROMs, and their evaluation does not require a computational mesh~\cite{regazzoni2024learning}.

In this context, however, dealing with geometric variability remains an open challenge. Surrogate models trained on a single geometry do not generalize across cohorts of differently-shaped anatomies, making it impractical to train one model per patient. At the same time, training a single model on a cohort of geometries is non-trivial in the absence of point-to-point correspondence between different anatomies, motivating the need for a geometric encoding of shape variability. Universal Solution Manifold Networks (USMNets), which we employ as starting point for the present work, in their base version achieve geometric encoding by means of landmark definitions \cite{regazzoni2022universal, zhang2025}, but do not straightforwardly generalize to complex geometries, where identifying landmarks is non-trivial. Other approaches use diffeomorphic mapping operators to map geometry variations onto a common reference space either to enhance deep learning based ROMs \cite{brivio2025handling} or to extract principal components of shape variability \cite{yin2024}. Applied to electrophysiology models of the left ventricle, the latter leverages universal ventricular coordinates (UVCs) \cite{bayer2018universal} to establish a mapping between patient-specific and reference geometries. However, both UVC fields and meshed representations are required, restricting applicability to structured data sources.
An alternative paradigm for encoding geometric variability is offered by representation learning techniques, which aim to describe complex data through a set of lower-dimensional features extracted in an automatic way \cite{bengio2013learning}.
In the context of shape modeling, this corresponds to discovering a latent space, where each point represents a geometry in the original three-dimensional domain, together with a mapping that allows reconstructing or querying the shape from this compact representation.
Deep autoencoder architectures \cite{dai2016shape, diederik2014auto-encoding} implement this idea by jointly training an encoder network that maps shapes to latent codes and a decoder network that reconstructs shapes from these codes. Related approaches, known as autodecoders, instead associate each shape with a learnable latent code optimized jointly with the decoder, removing the need for an explicit encoder and simplifying the overall architecture \cite{bojanowski2017optimizing}.
DeepSDF \cite{park2019deepsdf} introduces an autodecoder architecture for the simultaneous reconstruction of multiple geometries: each shape is associated with a low-dimensional latent code that conditions the network input and is optimized jointly with the network weights. The framework is flexible with respect to input data type, operating directly on point clouds, voxels, or meshes, and an explicit mesh representation could be extracted via the Marching Cubes algorithm \cite{lorensen1987marching}. Smoothness and regularity of the latent space can be further enforced by controlling the Lipschitz constant of the network \cite{liu2022learning}, promoting stable interpolation between geometries. The combination of DeepSDF with Lipschitz regularization has been successfully applied to reconstruct cardiac geometries in \cite{verhulsdonk24a}.

However, even when a geometric encoding method is available, learning the map from high-dimensional shape variability to the corresponding PDE solution remains a highly complex regression task. Approaches based on graph neural networks \cite{sanchezgonzalez2020learning-physics-graph, pfaff2021learning-mesh-graph} or attention mechanisms \cite{serrano2024aroma} typically require large training datasets to achieve robust generalization, and performance on unseen geometries can degrade significantly, when only limited training samples are available \cite{lanyon2025weaving}. This is an important limitation in biomedical applications, where open-access data is often scarce and acquisition of new high-fidelity simulations is computationally expensive. 

Motivated by these considerations, in this work, we present a method for building shape-informed surrogate models for cardiac mechanics from scattered point cloud data associated with a cohort of different patient anatomies.
We validate this framework on two datasets: an synthetic cohort of idealized geometries and two cohorts of real left ventricular anatomies from publicly available sources \cite{strocchi2020, rodero2021}.
The proposed pipeline consists of two steps. First a shape model is fitted to the geometric data, automatically learning a compact representation of each anatomy. When working with real data, the shape model is also used to generate new realistic geometries for data augmentation. Second, a shape-informed surrogate model is trained for predicting cardiac displacement by conditioning the network on the learned geometric encoding with the augmented dataset used to improve generalization.
The proposed framework, by decoupling the geometric complexity from the regression task, allows in this manner for data augmentation to compensate for the scarcity of biomedical training data.

For the shape modeling step, we propose two complementary approaches.
The first, named $\PCAmodel$ (PCA-based Shape Model), is based on Principal Component Analysis (PCA). Point-to-point correspondence between training geometries is established via UVCs, enabling PCA of the shape deformations. At inference time, the latent representation of a new geometry is obtained by finding the low-dimensional code, whose projection onto the PCA modes best approximates the ground truth.
The second, named $\SDFmodel$ (SDF-based Shape Model), is a neural network-based shape model that learns an implicit representation of each geometry by approximating its signed distance function (SDF), an approach commonly used in computer vision tasks \cite{xie2022neural}. This technique is extended to multiple geometries by adopting the DeepSDF autodecoder architecture \cite{park2019deepsdf}: a decoder $\NNsdf$ is conditioned on a shape code $\shapecode \in \mathcal{Z}$ and a point $\point$ sampled from the geometry, with optional additional geometric parameters $\geoparam$; the output is the SDF value at $\point$ for the geometry associated with $\shapecode$ and $\geoparam$. The latent distribution $\mathcal{Z}$ and the decoder weights are optimized jointly during training, combining a reconstruction loss, a prior on $\mathcal{Z}$, and a Lipschitz regularization term promoting smoothness of the latent space \cite{liu2022learning}. At inference time, the decoder weights are frozen and $\mathcal{Z}$ acts as a posterior distribution, enabling robust latent code optimization even for noisy or sparsely sampled point clouds.
To augment the dataset of real anatomies, $\SDFmodel$ is employed as a generative model: new realistic geometries are obtained by sampling the posterior distribution $\mathcal{Z}$ and conditioning the decoder on the sampled latent codes.
The second step consists in training a shape-informed surrogate model for cardiac displacement conditioned on the geometric encoding learned in the first step. The surrogate model is implemented as a neural network $\NNphysics$, optimized to predict the displacement field from the geometric encoding and the physical parameters of interest. We consider two variants depending on the source of the geometric encoding: $\PCAusm$, which uses the latent representation from $\PCAmodel$, and $\SDFusm$, which uses the latent code learned by $\SDFmodel$. Both variants are applied to the problem of predicting the displacement field for a finite strain mechanics problem of the left ventricle under diastolic loading conditions.

\begin{figure}[!htp]
\centering
\includegraphics[width=1\textwidth]{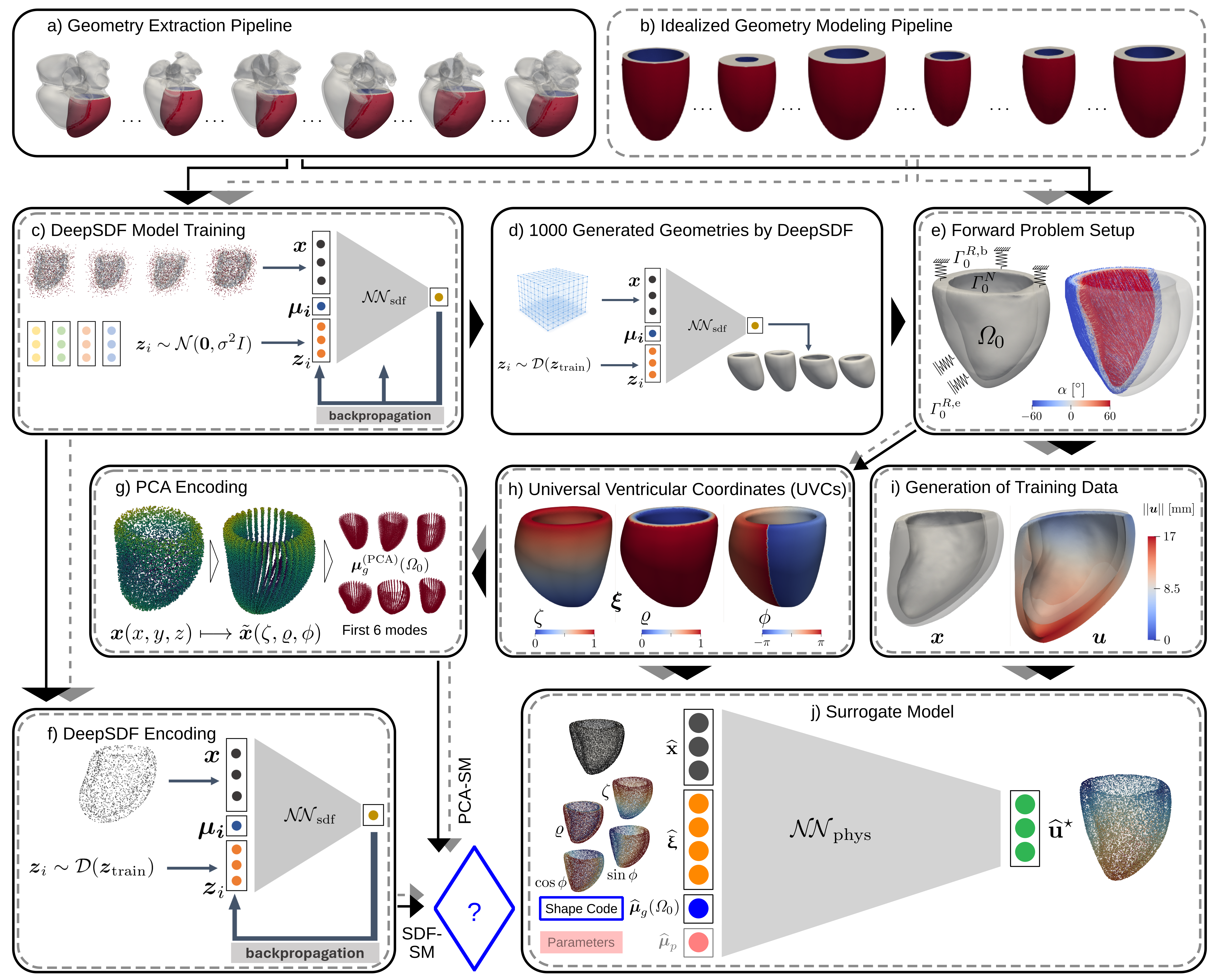}
\caption{Data processing, modeling, and training pipelines. Solid black lines/boxes show the pipeline of patient-specific left ventricular (LV) models; dashed gray lines/boxes that of the idealized LV models. They differ in that the idealized model cohort is not augmented by synthetically generated geometries.  a) LV geometry extraction from 4-chamber patient-specific heart geometries, retrieving 44 LV geometries.  b) Idealized LV models generated from varying long axis, diameter, and wall thickness $\ell$, $d$, and $w$, respectively, retrieving 512 models.  c) $\SDFmodel$ training: learning of signed-distance function to represent the geometry boundary.  d) Geometry generation by $\SDFmodel$.  e) Mechanical forward problem setup. Left: Reference geometry $\mathit{\Omega}_0$, with Robin (spring) boundaries on the base ($\mathit{\Gamma}_0^{R,\mathrm{b}}$) and the epicardium ($\mathit{\Gamma}_0^{R,\mathrm{e}}$) as well as a Neumann boundary at the endocardium ($\mathit{\Gamma}_0^{N}$). Right: Fiber field $\boldsymbol{f}_0$; color indicates the fiber angle $\alpha$ with respect to the circumference.  f) $\SDFmodel$ encoding.  g) $\PCAmodel$ encoding.  h) Computation of universal ventricular coordinates (UVCs).  i) Simulation of the forward model to generate displacement training data. Left: undeformed geometry. Right: deformed model with displacement field $\boldsymbol{u}$, color indicating its magnitude.  j) Surrogate model, informed by cartesian coordinates $\point$, UVCs $\boldsymbol{\upxi}$, a shape code $\boldsymbol{\mu}_{g}$ (either from $\PCAmodel$, $\SDFmodel$, or an analytic descriptor of shape features, e.g. features from the idealized geometries), and possibly some parameters $\boldsymbol{\mu}_{p}$.}\label{fig:pipeline}
\end{figure}

    \section{Results}
    \label{sec:results}

    We numerically test the proposed pipelines on two datasets: an idealized dataset and real patient anatomies. We first present the results on the shape encoding and reconstruction task (see Section~\ref{subsec:res_shape_reconstruction}) and then the performances on the surrogate modeling task (see Section~\ref{subsec:res_pde_inference}). 
The idealized dataset comprises 512 prolate ellipsoids parametrized by the triplet $(\ell, d, w)$, corresponding to long-axis length, diameter, and wall thickness, respectively, sampled on a uniform $8\times8\times8$ grid. This controlled dataset allows us to assess the shape model's ability to learn a descriptive latent space in a setting where the true parametrization is known analytically. In Section~\ref{subsubsec:idealized_geometries}, we analyze the correspondence between the learned low-dimensional representation and the ground-truth parameters, while Section~\ref{subsubsec:surrogate_idealized} compares surrogate model performance when conditioned on learned latent codes versus true geometric parameters.
Then, we consider a dataset of real patient anatomies consisting of 44 left ventricular geometries obtained from two publicly available cohorts: 24 heart failure patients \cite{strocchi2020} and 20 healthy patients \cite{rodero2021}. The complete data acquisition and preprocessing pipeline is described in Section~\ref{subsec:data_preprocessing}. On this dataset, we compare two shape encoding approaches, PCA- and SDF-based (respectively $\PCAmodel$ and $\SDFmodel$), for geometric representation (Section~\ref{subsubsec:real_geometries}) and for conditioning the surrogate model (PCA-USMNet and SDF-USMNet, respectively) (Section~\ref{subsub:surrogate_real}). The use of the $\SDFmodel$ for generating synthetic geometries for data augmentation is discussed in Section~\ref{subsec:res_shape_generation}. 

\subsection{Shape reconstruction}
\label{subsec:res_shape_reconstruction}
The quality of reconstructed geometries is assessed using two classes of metrics. Chamfer Distance ($\CD$) and Normalized Chamfer Distance ($\CDnorm$) measure the discrepancy between the reconstructed surface and the ground truth, considering the original and normalized geometry sizes, respectively. Intersection over Union ($\IoU$) and Dice Coefficient ($\Dice$) provide a volumetric measure of the overlap between the reconstructed shape and the ground truth, both in the $[0, 1]$ range, where higher values indicate better reconstruction. For $\SDFmodel$ we also report the Mean Level Estimate ($\MLE$) defined as the mean absolute SDF prediction on ground-truth surface points, where the correct value is zero by definition. A formal definition of all metrics is provided in the Methods (Section~\ref{subsec:shape_reconstruction_metrics}).

\subsubsection{Idealized geometries}
\label{subsubsec:idealized_geometries}

The reconstruction performance of $\SDFmodel$ is first evaluated on an idealized dataset. Since the parametrization is known, we also consider the case in which the network is conditioned directly on the true parameters rather than learning the latent space, and only the decoder weights are optimized.
This serves as a best-case comparison to decouple the representation capability of the decoder and the descriptive power of the latent space. The input to the network includes also the scaling factor $\mu_i$ associated to each shape $\Omega_i$. Since this scaling factor relates strongly to $\ell$ (see Methods Section~\ref{subsec:dataset_generation}), we consider it as the third component of the latent code, without learning it.

Table~\ref{tab:reconstruction_metrics_idealized} presents the reconstruction metrics for both approaches. The results indicate no significant difference in reconstruction quality between the learned latent codes and the original parametrization, suggesting that the network successfully captures the essential geometric information in its learned representation.

\begin{table}[htbp]
\centering
\caption{Reconstruction metrics for $\SDFmodel$ trained with learned latent codes versus original parametric features on idealized geometries.}
\label{tab:reconstruction_metrics_idealized}
\begin{tabular}{llccccc}
\toprule
\textbf{Code}  & \textbf{Split} & \textbf{$\CD$} \textbf{[mm]} & \textbf{$\CDnorm$} & \textbf{$\IoU$} & \textbf{$\Dice$} & \textbf{$\MLE$}\\
\midrule
\multirow{2}{*}{Learned} &
Test & $ 1.276 \pm 0.282 $ & $ 0.0238 \pm 0.0062$ & $ 0.953 \pm 0.027$ & $ 0.976 \pm 0.014$ & $ 1.367 \pm 0.587$ \\
& Train & $ 1.232 \pm 0.275$ & $0.0226 \pm 0.0059$ & $ 0.954 \pm  0.022$ & $0.976 \pm 0.012$ & $ 1.257\pm 0.614$ \\
\midrule
\multirow{2}{*}{Original} &
Test & $ 1.124\pm  0.290$ & $0.0209 \pm 0.0040$ & $0.950 \pm 0.036$ & $0.974 \pm 0.020$ & $1.381 \pm 0.596$ \\
& Train & $1.100 \pm 0.133$ & $0.0203 \pm 0.0036$ & $0.954 \pm 0.026$ & $0.976 \pm 0.014$ & $1.261 \pm 0.614$\\
\bottomrule
\end{tabular}
\end{table}

\paragraph{Latent space structure and interpretability.}

\begin{figure}[!htp]
\centering
\includegraphics[trim= 0 13 0 24,clip]{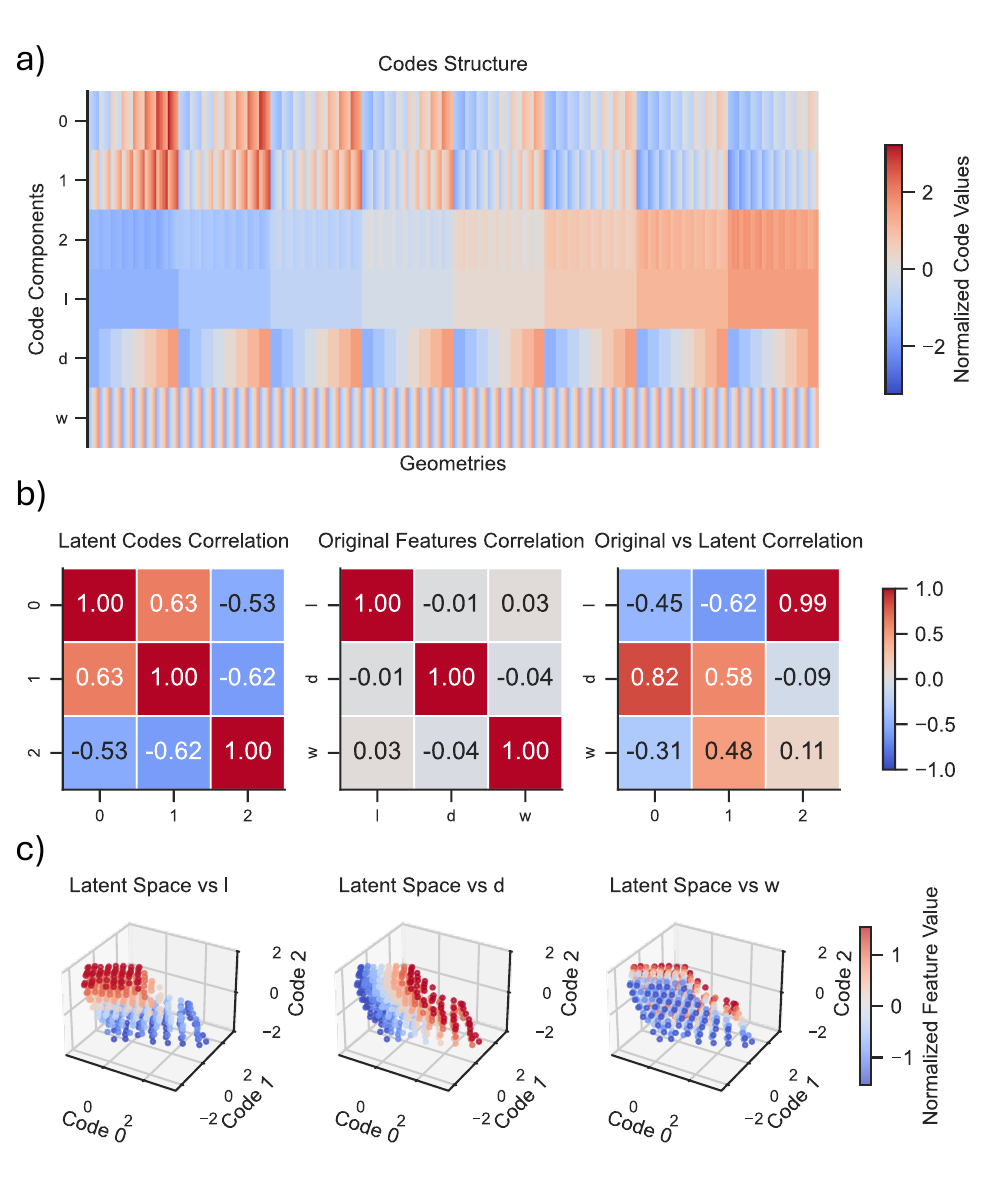}
\caption{Visualizations of the learned latent space for the idealized geometries dataset. a) Comparison between reconstructed latent codes and original geometric features for the idealized geometries. Each column corresponds to an individual geometry, with colors representing the z-normalized value of the associated original feature to ensure homogeneous color scaling across dimensions. b) Pairwise Pearson correlation matrices between the learned latent codes and the original geometric features for the idealized geometries. 
c) Three-dimensional visualization of the learned latent space for the idealized geometries. Points represent individual samples in the space of the first three latent codes and are colored according to the z-normalized values of the original geometric features, illustrating how variations in physical parameters are embedded in the latent representation.}\label{fig:idealized_overall}
\end{figure}

To investigate the relationship between learned latent codes and the original geometric parameters, we analyze the structure of the learned representation.
Figure~\ref{fig:idealized_overall}a displays the component values of both learned and original codes across all 512 geometries. The $8 \times 8 \times 8$ grid structure used to generate the original parameter triplets is clearly visible in both representations. We observe that $\mu_i$, here indicated as the third component of the latent code, is strongly correlated with the original feature $\ell$. Figure~\ref{fig:idealized_overall}b shows the pairwise Pearson    correlation matrices. While the learned codes capture relationships present in the original features, they are not independent.
The three-dimensional distribution of the learned codes is shown in Figure~\ref{fig:idealized_overall}c.

To quantify the relationship between learned codes and original features, we fit linear regression models to predict each original parameter from the learned latent codes. The models are trained on codes from the training set and evaluated on the test set. 
The linear regression models achieve high $R^2$ values both on the test set ($R^2_\ell = 0.988$, $R^2_d = 0.941$, $R^2_w = 0.965$) and the train set ($R^2_\ell = 0.995$, $R^2_d = 0.929$, $R^2_w = 0.957$), indicating that the original geometric parameters are recoverable from the latent representation through a linear mapping.

\subsubsection{Real Geometries}
\label{subsubsec:real_geometries}

\paragraph{Reconstruction results}

\begin{figure}[!htp]
\centering
\includegraphics{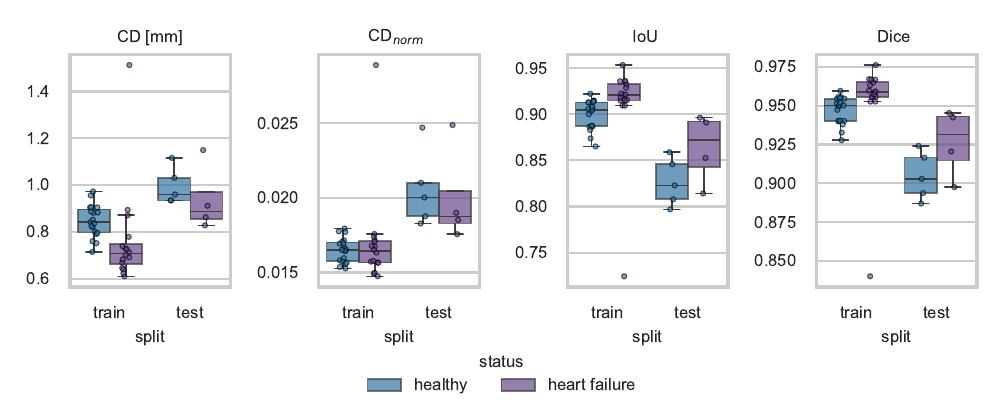}
\caption{Performance metrics for $\SDFmodel$ reconstructed geometries.}\label{fig:evaluation_metrics_model}
\end{figure}

The model achieves a reconstruction error below 1 mm in terms of CD both on the training and test set. We also observe that there is no significant difference in $\CDnorm$ between the two cohorts of patients, as the size difference is effectively factored out in the preprocessing stage in Alg.~\ref{algo:mesh_preprocessing}. In terms of intersection over union, the worst performance on the test set is $\IoU_{\text{min}}=0.797$. The complete reconstruction performances are reported in Table~\ref{tab:reconstruction_metrics_compact_model} and in Figure~\ref{fig:evaluation_metrics_model}.

\begin{table}[htbp]
\centering
\caption{Reconstruction metrics for $\SDFmodel$ by cohort and data split (mean ± std).}
\label{tab:reconstruction_metrics_compact_model}
\begin{tabular}{llcccc}
\toprule
\textbf{Cohort} & \textbf{Split} & \textbf{$\CD$} \textbf{[mm]} & \textbf{$\CDnorm$} & \textbf{$\IoU$} & \textbf{$\Dice$} \\
\midrule
\multirow{2}{*}{Healthy} 
  & Test  & $0.994 \pm 0.078$ & $0.0205 \pm 0.0026$ & $0.826 \pm 0.026$ & $0.904 \pm 0.015$ \\
  & Train & $0.846 \pm 0.069$ & $0.0165 \pm 0.0008$ & $0.899 \pm 0.016$ & $0.947 \pm 0.008$ \\
\midrule
\multirow{2}{*}{Heart Failure} 
  & Test  & $0.938 \pm 0.145$ & $0.0200 \pm 0.0033$ & $0.864 \pm 0.038$ & $0.926 \pm 0.022$ \\
  & Train & $0.763 \pm 0.215$ & $0.0170 \pm 0.0033$ & $0.912 \pm 0.051$ & $0.953 \pm 0.031$ \\
\bottomrule
\end{tabular}
\end{table}

\begin{figure}[!htp]
\centering
\includegraphics[trim= 0 80 0 10,clip]{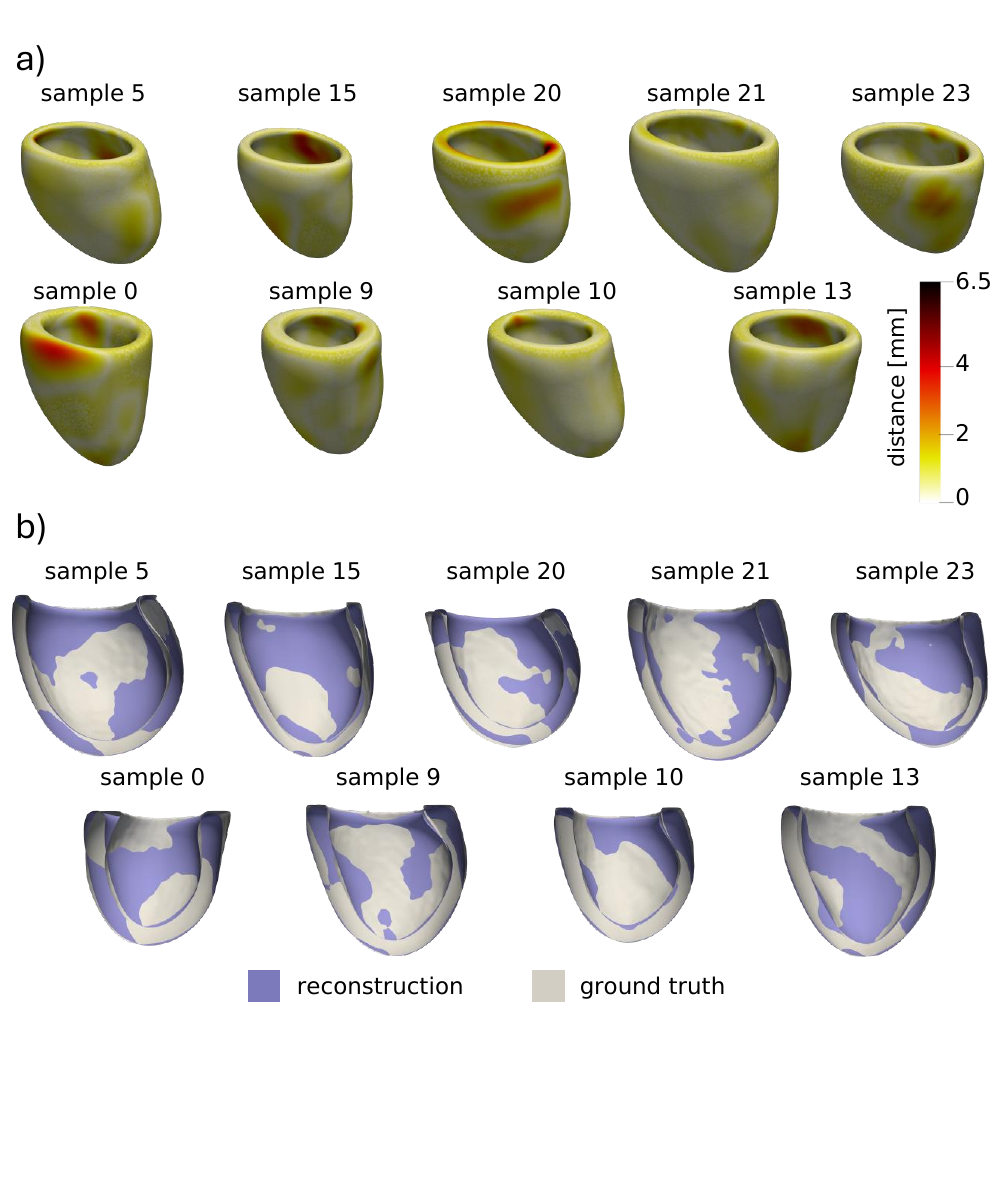}
\caption{a) Reconstructed meshes colored according to the distance [mm] to the closest point on the target mesh. First row: heart failure patients, second row: healthy patients. b) Reconstructed meshes (in purple) and original ones (in white). First row: heart failure patients; second row: healthy patients.}\label{fig:full_rec} 
\end{figure}

Figure~\ref{fig:full_rec}b compares the section of the reconstructed test geometries with the original ones, while Figure~\ref{fig:full_rec}a provides a three-dimensional visualization. 

\paragraph{Latent space analysis}

\begin{figure}[!htp]
\centering
\includegraphics[trim= 0 90 0 20,clip]{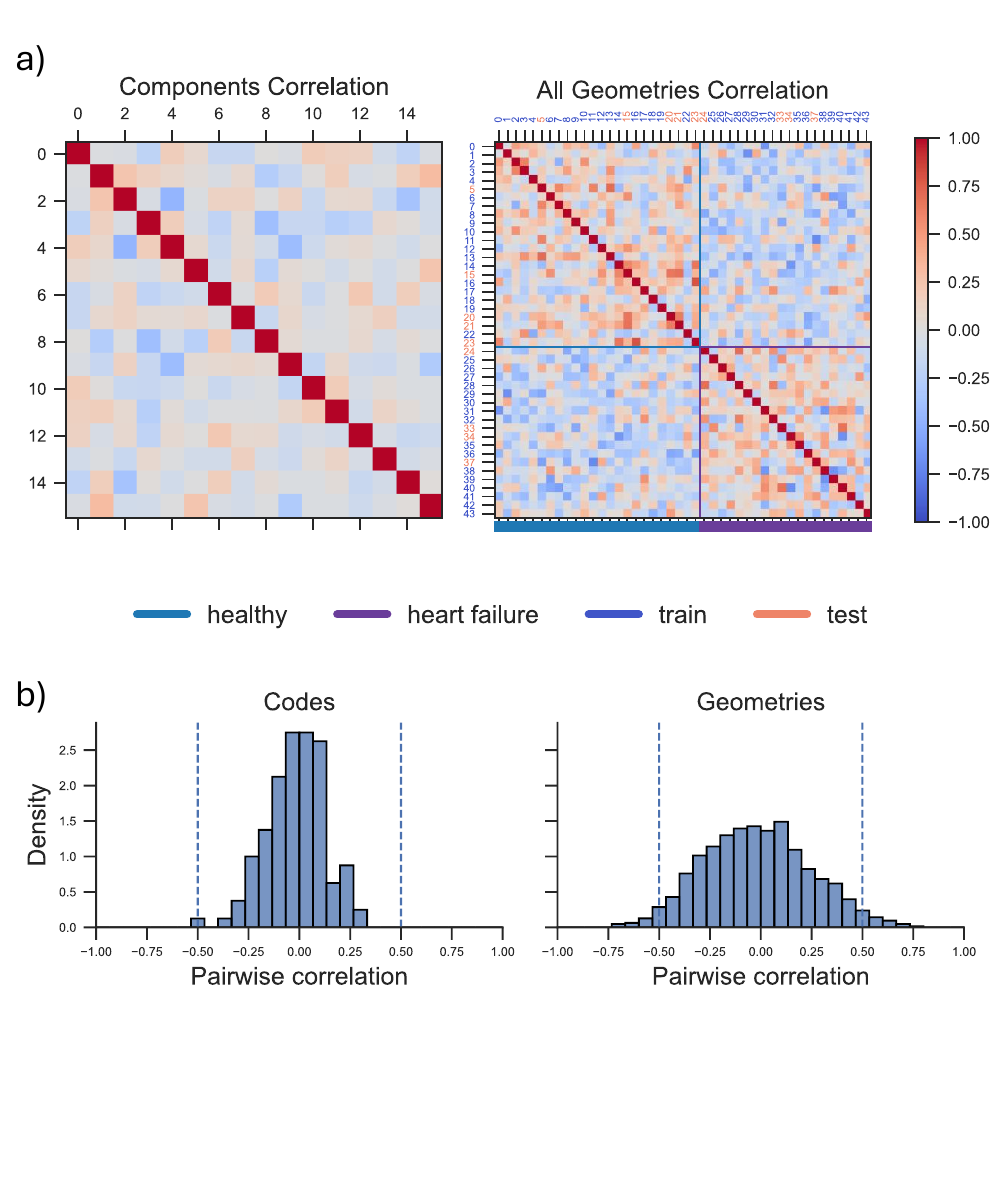}
\caption{a) Correlation matrices computed after inference. Matrix on the left describes correlation among different components of the latent space. Matrix on the right describes correlation between codes of different geometries. Colored line at the bottom distinguishes the two cohorts of patients, while the color of the index indicates wether the geometry belongs to training or test set. b) Empirical distributions of pairwise correlations respectively between codes and geometries. Histograms show the density of the upper triangular elements of the corresponding correlation matrices (excluding the diagonal), so that each variable pair is counted once. Vertical dashed lines indicate correlation thresholds of $\pm 0.5$.}\label{fig:correlation_analysis}
\end{figure}

Figure~\ref{fig:correlation_analysis} shows the correlation matrices and distributions among the empirical latent space components and among the latent representations of the 44 patients anatomies. 

We observe that the empirical latent space shows weak pairwise correlation (range: $\rho_{\text{min}}=-0.501$ to $\rho_{\text{max}}=0.3079$), indicating that the 16-dimensional latent representation captures largely independent geometric features. 
In contrast, more significant correlation patterns emerge among latent representations of patient anatomies, especially within cohorts. Anatomies from the same patient cohort exhibit more pronounced correlations ($\rho_{\text{min}}=-0.699$ and $\rho_{\text{max}}=0.808$), which suggests that the latent space clusters geometries depending on some cohort specific anatomical characteristics.

This structure shows that the learned representation successfully captures both inter-patient and cohort-specific variability, which can be beneficial for downstream surrogate model conditioning based on anatomical features.

\paragraph{PCA reconstruction}

Figure~\ref{fig:pca_modes} in Appendix~\ref{sec:additional_materials} shows the average shape obtained by projecting each geometry onto a uniform UVC grid, together with the deviations generated by adding the first 6 modes obtained through PCA. 

\begin{table}[htbp]
\centering
\caption{Reconstruction metrics for $\PCAmodel$ by cohort and data split (mean ± std).}
\label{tab:reconstruction_metrics_compact_pca}
\begin{tabular}{llcc}
\toprule
\textbf{Cohort} & \textbf{Split} & \textbf{$\CD$} \textbf{[mm]} & \textbf{$\CDnorm$} \\
\midrule
\multirow{2}{*}{Healthy} 
  & Test  & $0.807 \pm 0.063$ & $0.0167 \pm 0.0021$ \\
  & Train & $0.567 \pm 0.129$ & $0.0111 \pm 0.0027$ \\
\midrule
\multirow{2}{*}{Heart Failure} 
  & Test  & $0.740 \pm 0.130$ & $0.0158 \pm 0.0030$ \\
  & Train & $0.484 \pm 0.080$ & $0.0109 \pm 0.0013$ \\
\bottomrule
\end{tabular}
\end{table}

\begin{figure}[!htp]
\centering
\includegraphics{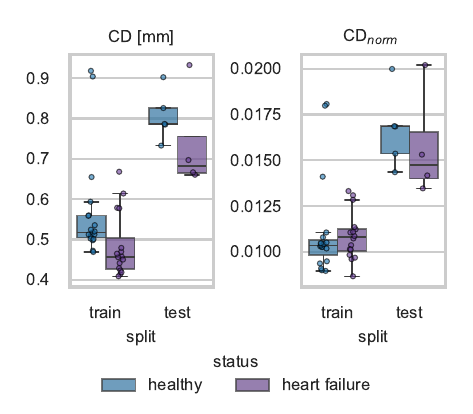}
\caption{Performance metrics for $\PCAmodel$ on reconstructed geometries.}\label{fig:pca_metrics}
\end{figure}

Figure~\ref{fig:pca_metrics} and Table~\ref{tab:reconstruction_metrics_compact_pca} report performance metrics for $\PCAmodel$. We observe that in this setting $\PCAmodel$ outperforms $\SDFmodel$ with respect to $\CD$ and $\CDnorm$, while volume-level metrics like $\IoU$ and $\Dice$ are not computed, since $\PCAmodel$ does not produce a closed watertight surface. 

\subsubsection{Robustness Evaluation (noise numerosity test and PCA comparison)}
\label{subsubsec:quality_evaluation}

The reconstruction performances of the $\SDFmodel$ and $\PCAmodel$ are evaluated with varying levels of noise and sampling. The sampling value varies between $\{125, 250, 500, 1000, 2000\}$ points, while the noise standard deviation in $\{0.0, 0.0125, 0.025, 0.05, 0.1\}$ is applied after normalizing the mesh, in order to perturb all elements of the cohort with the same relative amount of noise. On average, the corresponding noise levels in the non-normalized space is $\{0.0, 0.6\,\text{mm}, 1.2\,\text{mm}, 2.4\,\text{mm}, 4.8\,\text{mm}\}$.

\begin{figure}[!htp]
\centering
\includegraphics{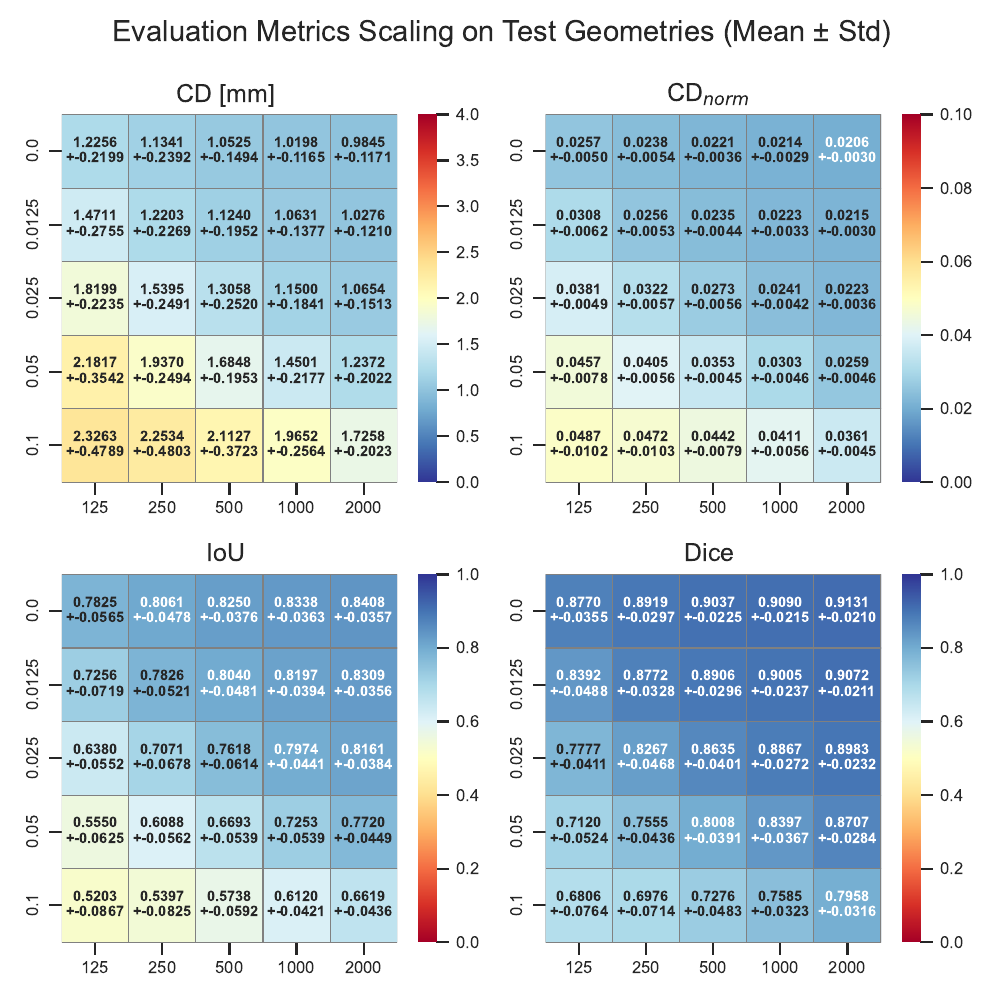}
\caption{$\SDFmodel$ performance on test geometries as a function of sample size and geometric noise level.}\label{fig:heatmap_model} 
\end{figure}

Figure~\ref{fig:heatmap_model} shows how the reconstruction metrics for $\SDFmodel$ vary with geometric perturbation level and sample size. As expected from the probabilistic inference formulation (Section~\ref{subsec:shape_model_inference}), reconstruction quality degrades with increasing noise but improves with larger sample sizes. This behavior reflects the adaptive regularization mechanism: the Mahalanobis prior weight $w_{\text{post}}$ decreases with sample size, allowing the optimization to rely more heavily on data when more observations are available, and more on the learned prior when data is scarce or noisy. The IoU remains above 0.5 even in the most challenging scenario (125 points with maximum noise level), demonstrating the robustness of the probabilistic latent code optimization.

\begin{figure}[!htp]
\centering
\includegraphics{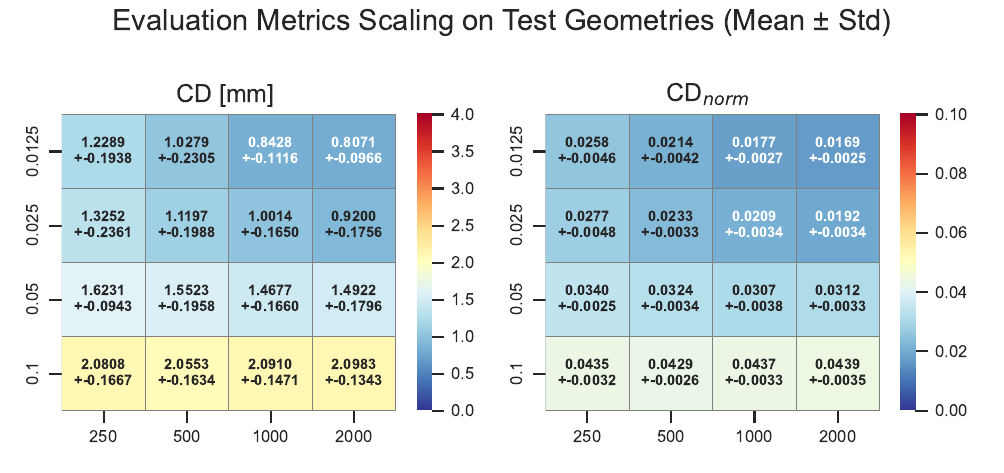}
\caption{$\PCAmodel$ performance on test geometries as a function of sample size and geometric noise level.}\label{fig:heatmap_pca} 
\end{figure}

Figure~\ref{fig:heatmap_pca} shows that $\PCAmodel$ performance degrades with increasing noise but exhibits minimal improvement when sample size increases. This reflects the difference in the reconstruction approach, where $\PCAmodel$ coefficients are found by minimizing $\CD$ between the linearly reconstructed point cloud and noisy observation, without explicitly modeling measurement noise and sample size.

For low noise levels ($\sigma \leq 0.025$) or small sample sizes ($N \leq 500$), $\PCAmodel$ achieves lower reconstruction error, benefiting from its compact linear parameterization of the shape space learned from the training cohort. On the other hand, $\SDFmodel$ provides better performances for high noise and higher sample size regime ($N > 500, \sigma >  0.05$), where the probabilistic inference framework is able to denoise the latent code estimate by giving more importance to the learned posterior distribution.

In order to describe with a single quantity the amount of information carried by the data under different noise and sampling conditions, we define the \textit{effective noise level} as
\begin{equation}
\label{eq:effective_noise_level}
\rho = \frac{\sigma_{\text{geo}}}{\sqrt{N_{\text{geo}}}},
\end{equation}
where $\sigma_{\text{geo}}$ is the standard deviation of the Gaussian noise applied to the point coordinates and $N_{\text{geo}}$ is the number of sampled points.
This scaling naturally arises from the statistical aggregation of independent measurements: while the variance of averaged noise decreases proportionally to $1/N$, the corresponding standard deviation scales as $1/\sqrt{N}$. The parameter $\rho$ thus quantifies the residual noise amplitude per effective degree of information.  

The effective noise level can also be interpreted as a measure of label error. Since signed distance functions satisfy the eikonal equation $\|\nabla \text{SDF}\| = 1$, they are 1-Lipschitz continuous: perturbing a point $\mathbf{x}$ by $\epsilon$ shifts the SDF value by at most $\|\epsilon\|_2$, as $|\text{SDF}(\mathbf{x}) - \text{SDF}(\mathbf{x} + \epsilon)| \leq \|\epsilon\|_2$. Consequently, geometric noise on the input coordinates with standard deviation $\sigma_{\text{geo}}$ induces a label error of similar magnitude. When aggregated over $N_{\text{geo}}$ independent noisy observations, the effective label uncertainty scales as $\sigma_{\text{geo}}/\sqrt{N_{\text{geo}}}$, which directly affects the quality of the latent code recovered at inference and ultimately the accuracy of the reconstructed zero-level set.

\begin{figure}[!htp]
\centering
\includegraphics{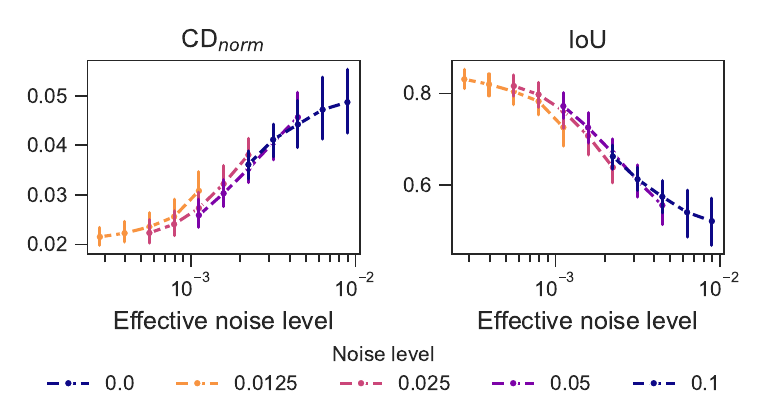}
\caption{$\CDnorm$ and $\IoU$ for $\SDFmodel$ as a function of effective noise level $\rho$ on test geometries. Each point corresponds to a different $(\sigma_{\text{geo}}, N_{\text{geo}})$ combination. Error bars represent one standard deviation.}\label{fig:noise_density_model} 
\end{figure}

This intuition is confirmed by Figure~\ref{fig:noise_density_model}, which shows that reconstruction errors obtained for different pairs $(\sigma, N)$ collapse onto the same curve when plotted against $\rho$. In particular, configurations sharing the same value of $\rho$ yield nearly identical accuracy, despite corresponding to different absolute noise levels and sampling densities. This confirms that $\rho$ is the relevant scaling parameter governing the reconstruction performance and provides practical guidance for experimental design: increasing the number of sampled points compensates for higher noise levels according to the predictable trade-off $\rho = \sigma_{\text{geo}}/\sqrt{N_{\text{geo}}}$.

\begin{figure}[!htp]
\centering
\includegraphics{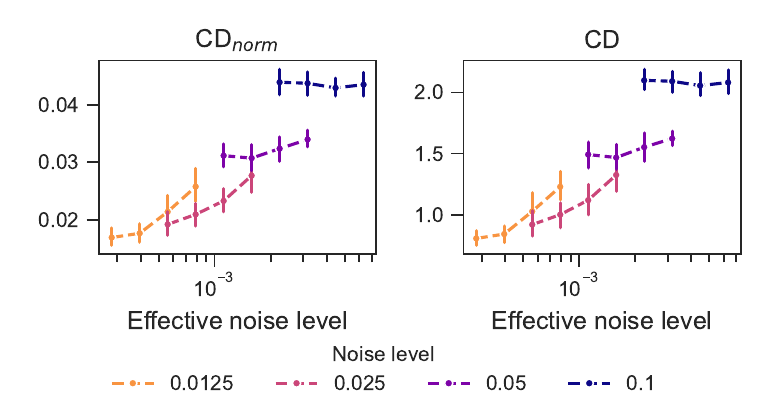}
\caption{$\CDnorm$ and $\CD$ for $\PCAmodel$ as a function of effective noise level $\rho$ on test geometries. Unlike $\SDFmodel$ (Fig.~\ref{fig:noise_density_model}), points do not collapse onto a single curve, indicating that $\rho$ does not capture the full behavior of the optimization. Error bars represent one standard deviation.}\label{fig:noise_density_pca} 
\end{figure}

In contrast, Figure~\ref{fig:noise_density_pca}  shows that $\rho$ does not characterize $\PCAmodel$ performance as effectively: points corresponding to different $(\sigma_{\text{geo}}, N_{\text{geo}})$ combinations do not collapse onto a single curve. The noise level $\sigma_{\text{geo}}$ is discriminant, while the sample size $N_{\text{geo}}$ does not impact the reconstruction quality in an equal manner.

\subsection{Synthetic geometries generation}
\label{subsec:res_shape_generation}
\begin{figure}[!htp]
\centering
\includegraphics[width=0.8\textwidth]{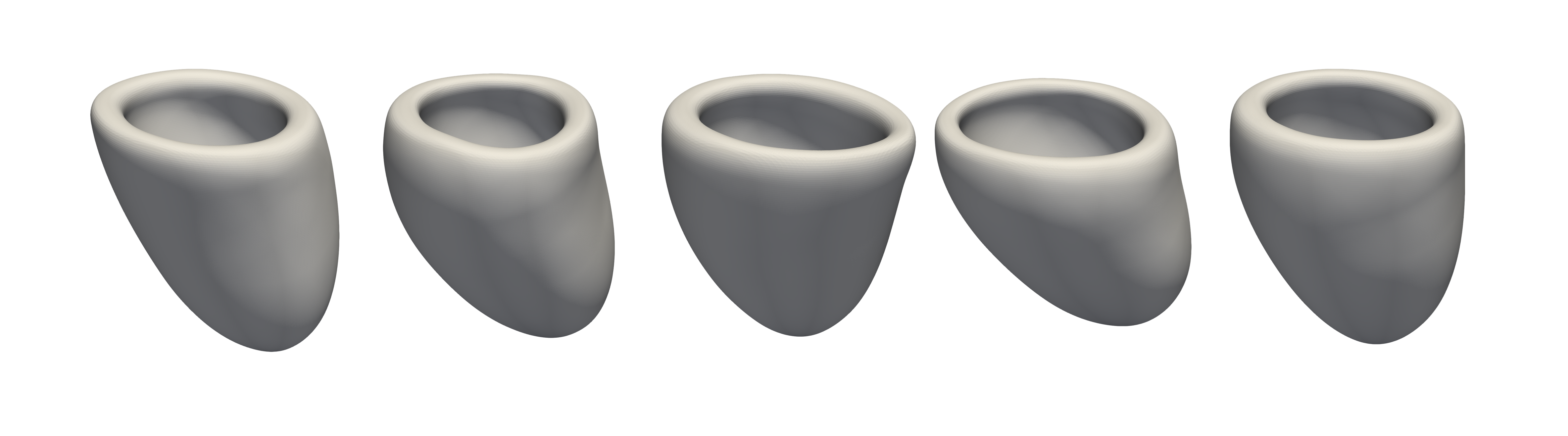}
\caption{Examples of five successfully generated synthetic left ventricular geometries.}
\label{fig:gen_geo5}
\end{figure}

We use the latent space and decoder combination learned by $\SDFmodel$ in order to generate new realistic geometries that can be used for data augmentation in the surrogate modeling task. Following the procedure described in Section~\ref{subsec:shape_generation}, we sample 1000 latent codes from the empirical distribution $\mathcal{N}(\boldsymbol{\mu}, \Sigma)$ and generate corresponding synthetic geometries. Of these, 976 geometries are anatomically plausible and retained for data augmentation, while 24 samples are discarded due to topological artifacts such as holes close to the apex or degeneration in the base surface. Figure~\ref{fig:gen_geo5} show examples of successfully generated geometries, while some discarded examples are reported in Figure~\ref{fig:broken5} in Appendix~\ref{sec:additional_materials}.

\subsection{Shape-dependent surrogate model}
\label{subsec:res_pde_inference}
In this section, we report results on training of the cardiac mechanics surrogate $\NNphysics$. We focus on quantifying the effect of shape encoding and dataset augmentation, as well as the influence of query points and loss augmentation by strains. First, we show results on the idealized cohort of artificially generated left ventricle geometries, and then investigate the real cohort of patient-specific geometries, as standalone dataset and as dataset augmented by synthetically generated models.

\subsubsection{Idealized geometries}
\label{subsubsec:surrogate_idealized}
\begin{figure}[!htp]
\vspace{-.2cm}
\centering
\includegraphics[width=0.68\textwidth]{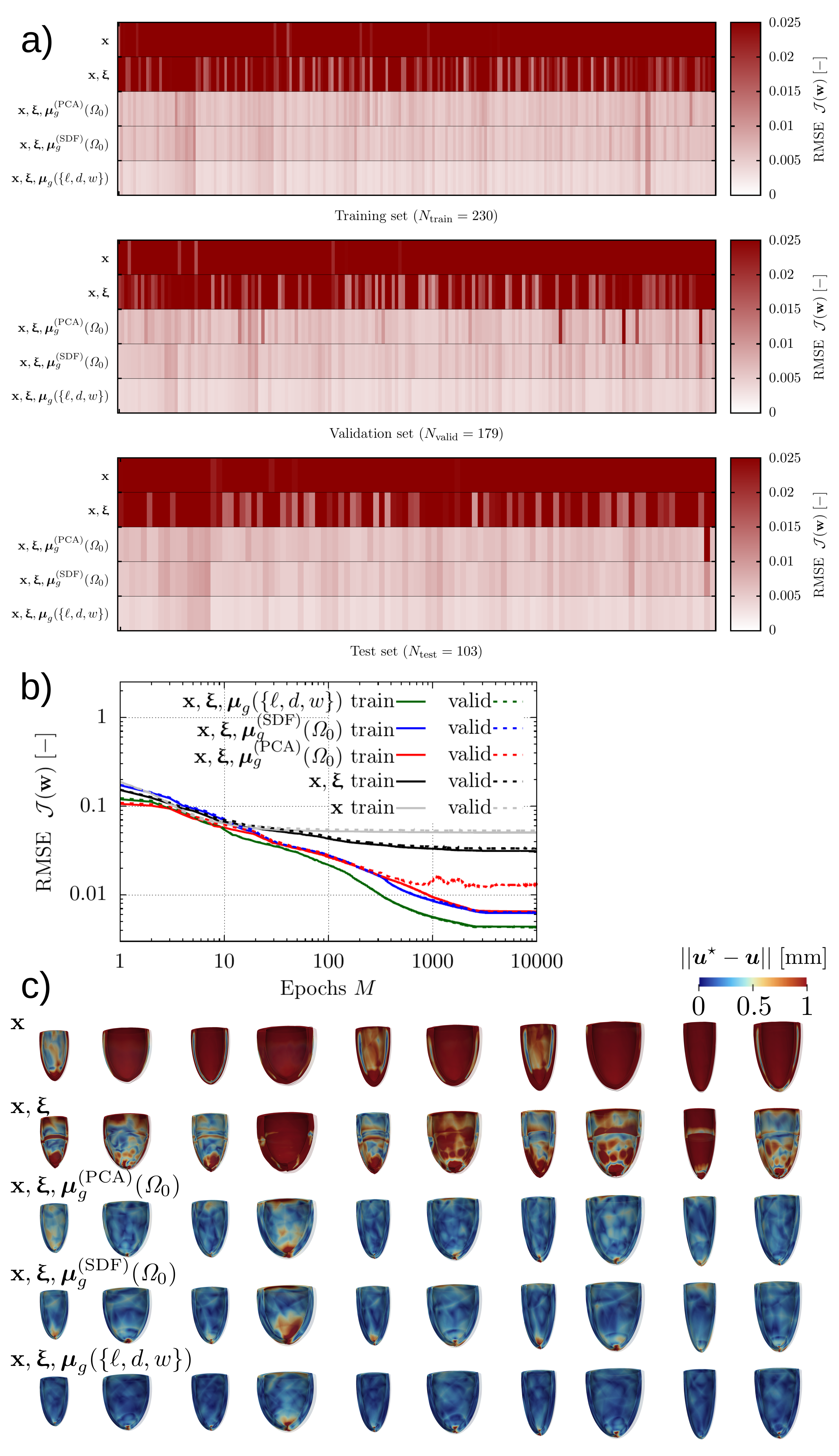}
\caption{Idealized geometries: Comparison of different input layers and shape codes ($\mathbf{x}$: only Cartesian coordinates; $\mathbf{x},\boldsymbol{\upxi}$: Cartesian coordinates plus UVCs; $\mathbf{x},\boldsymbol{\upxi},\boldsymbol{\mu}_{g}^{(\mathrm{PCA})}(\mathit{\Omega}_0)$: Cartesian coordinates, UVCs, $\PCAmodel$ shape code; $\mathbf{x},\boldsymbol{\upxi},\boldsymbol{\mu}_{g}^{(\mathrm{SDF})}(\mathit{\Omega}_0)$: Cartesian coordinates, UVCs, $\SDFmodel$ shape code; $\mathbf{x},\boldsymbol{\upxi},\boldsymbol{\mu}_{g}(\{\ell,d,w\})$: Cartesian coordinates, UVCs, analytic shape code). No strain contributions to the loss function ($\lambda_{\mathrm{s}}=0$), number of query points $n=15000$. a) RMSE on training, validation, and test set for the five different input layers. b) RMSE over training epochs, split into training and validation loss. c) Absolute difference in ground-truth and inferred displacement on ten selected test geometries.}\label{fig:results_ideal_sc}
\end{figure}

\begin{figure}[!htp]
\centering
\includegraphics[width=1\textwidth]{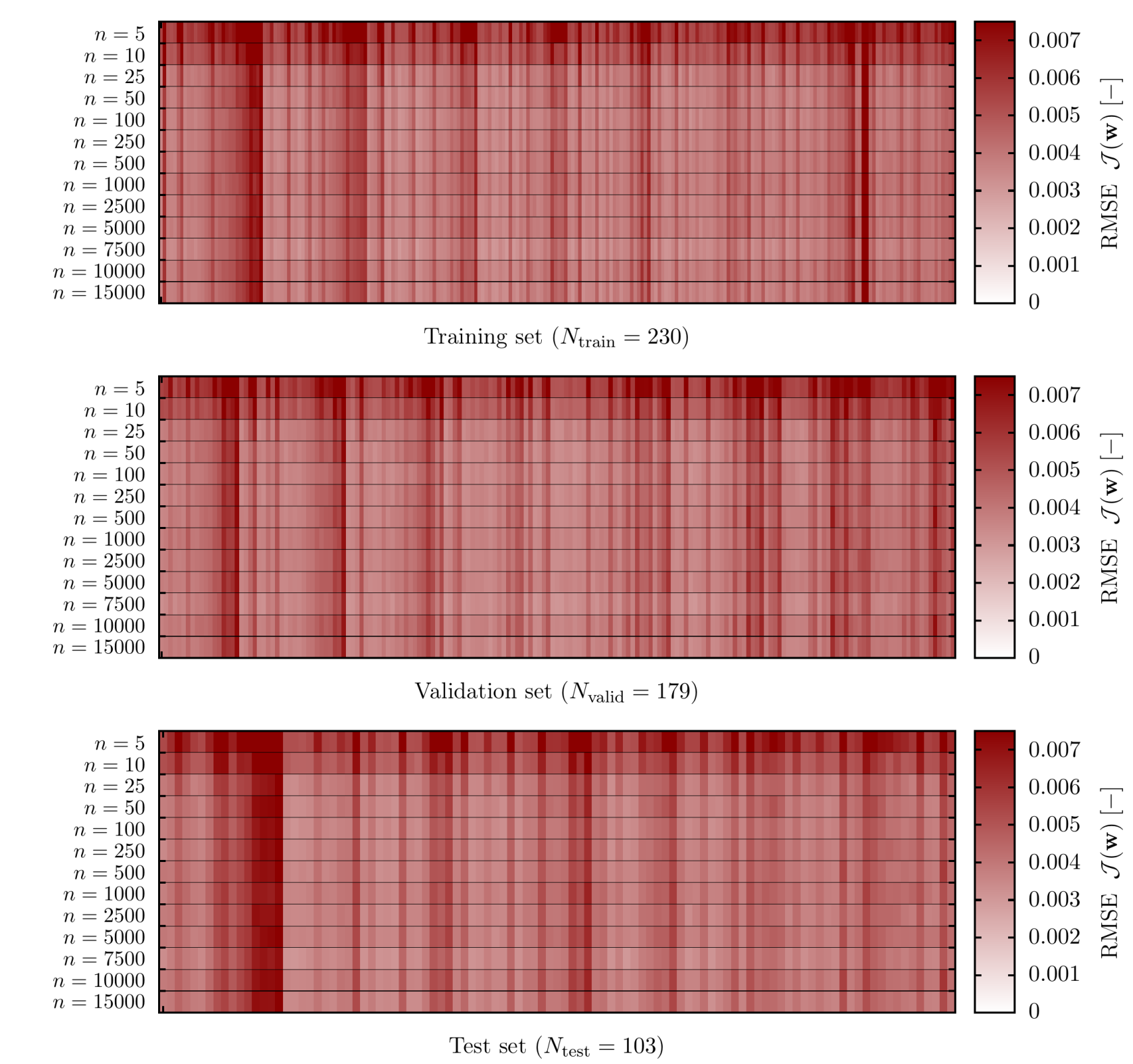}
\caption{Idealized geometries: RMSE on training, validation, and test set for different number of query points, $n\in[5,\hdots,15000]$. No strain contributions to the loss function ($\lambda_{\mathrm{s}}=0$).}\label{fig:results_ideal_points}
\end{figure}

\begin{figure}[!htp]
\centering
\includegraphics[width=0.75\textwidth]{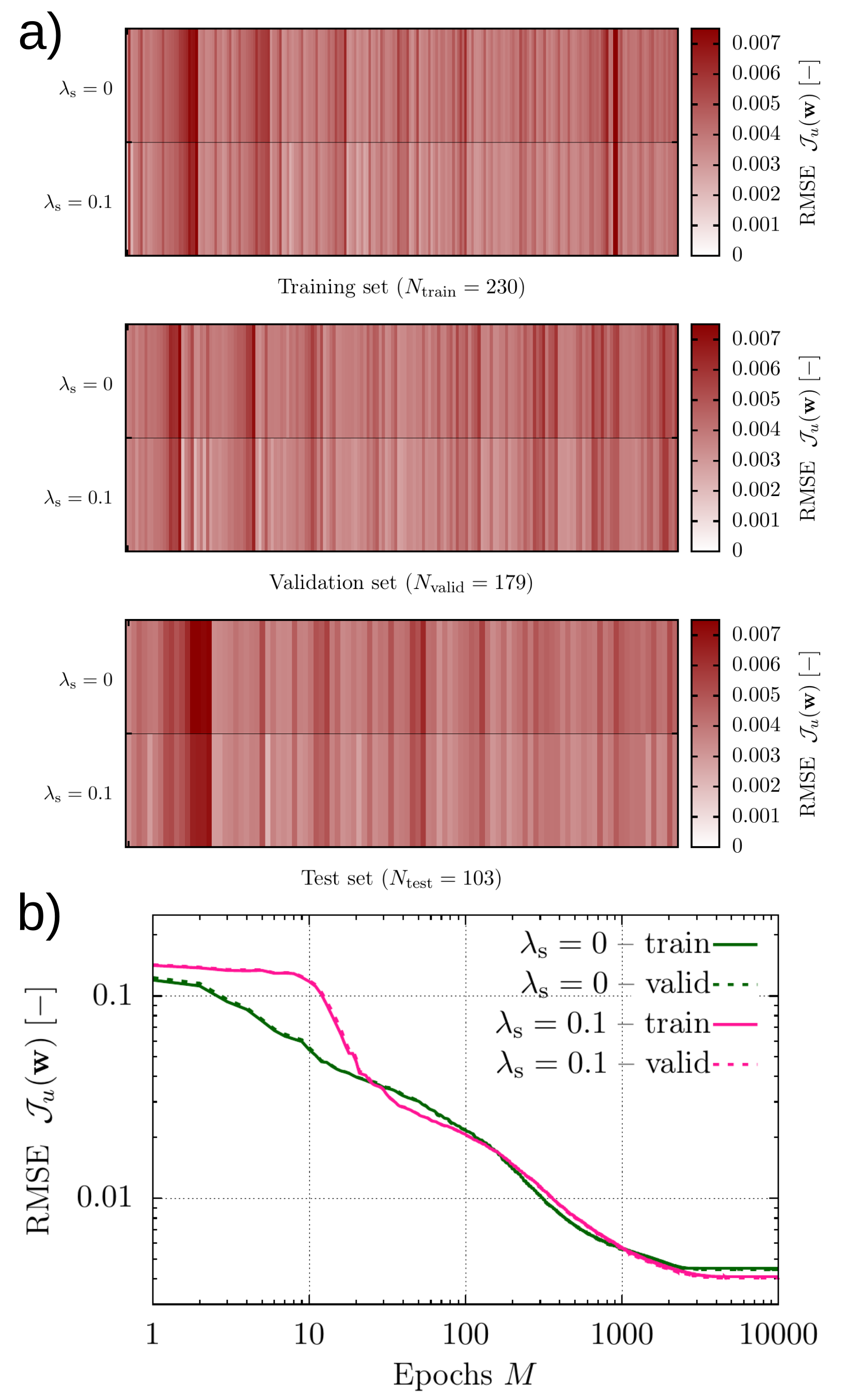}
\caption{Idealized geometries, investigation of effects of strain contributions to the loss function. No strain ($\lambda_{\mathrm{s}}=0$) vs. strain loss, with $\lambda_{\mathrm{s}}=0.1$ chosen such that both loss contributions are of comparable magnitude; number of query points $n=2500$. a) Displacement RMSE on training, validation, and test set. b) Displacement RMSE over number of training epochs, split into training and validation loss.}\label{fig:results_ideal_strain}
\end{figure}

We train the surrogate model with displacement solutions of the cohort of 512 prolate ellipsoid geometries (Figure~\ref{fig:pipeline}b) and investigate the influence on the root mean squared error (RMSE) of: shape encoding and landmarking (Figure~\ref{fig:results_ideal_sc}); the number of query points (Figure~\ref{fig:results_ideal_points}); and the presence of strain information (Figure~\ref{fig:results_ideal_strain}).

\paragraph{Shape encoding and landmarking}
To test the influence of shape codes and geometric landmarks (e.g., UVCs), we train our neural network using $n=15000$ randomly selected points from the geometry, and only consider displacement information in Eq.~(\ref{eq:Jw}), hence $\lambda_{\mathrm{s}}=0$ (no strain-based regularization). For efficiency purposes, we bypass any strain calculation.

Figure~\ref{fig:results_ideal_sc}a shows the RMSE on the training ($N_{\mathrm{train}}=230$), validation ($N_{\mathrm{valid}}=179$), and test set ($N_{\mathrm{test}}=103$), Figure~\ref{fig:results_ideal_sc}b the RMSE over the number of training epochs (split into training and validation error), and Figure~\ref{fig:results_ideal_sc}c the plotted absolute difference in ground-truth and inferred displacement for ten selected geometries from the test set. Five different shape encodings/geometric landmarkings are used:
\begin{itemize}
    \item $\mathbf{x}$: Input layer only consisting of Cartesian coordinates;
    \item $\mathbf{x},\boldsymbol{\upxi}$: Input layer consisting of Cartesian coordinates and UVCs;
    \item $\mathbf{x},\boldsymbol{\upxi},\boldsymbol{\mu}_{g}^{(\mathrm{SDF})}(\mathit{\Omega}_0)$: Input layer consisting of Cartesian coordinates, UVCs, and a DeepSDF shape code of dimension 3;
    \item $\mathbf{x},\boldsymbol{\upxi},\boldsymbol{\mu}_{g}^{(\mathrm{PCA})}(\mathit{\Omega}_0)$: Input layer consisting of Cartesian coordinates, UVCs, and a PCA shape code of dimension 6;
    \item $\mathbf{x},\boldsymbol{\upxi},\boldsymbol{\mu}_{g}(\{\ell,d,w\})$: Input layer consisting of Cartesian coordinates, UVCs, and a set of length, diameter, and wall thickness of the idealized shape.
\end{itemize}
The first column block of Table~\ref{tab:errors_all_shapecode} shows the averaged errors over the training, validation, and test sets.
Compared to the na\"ive setting providing only $\mathbf{x}$ to the input layer, adding UVCs, $\boldsymbol{\upxi}$, alone already leads to a 40.2\% (training set), 39.9\% (validation set), and 40.6\% (test set) improvement in RMSE, respectively. Providing $\PCAmodel$, $\boldsymbol{\mu}_{g}^{(\mathrm{PCA})}(\mathit{\Omega}_0)$, or $\SDFmodel$, $\boldsymbol{\mu}_{g}^{(\mathrm{SDF})}(\mathit{\Omega}_0)$, shape codes then improves the error further by 77.7\% and 78.7\% (training set), 73.3\% and 79.2\% (validation set), and 73.4\% and 77.4\% (test set), respectively. Only informing the neural network by the analytic idealized model parameterization $\boldsymbol{\mu}_{g}(\{\ell,d,w\})$ can achieve an even more substantiated error reduction: by 85.1\% (training set), 85.8\% (validation set) and 84.5\% (test set) compared to using $\mathbf{x},\boldsymbol{\upxi}$ only.

\paragraph{Number of query points}
We investigate how the number of (randomly subsampled) query points per geometry used in training affects the losses evaluated on the entirety of geometric points. Figure~\ref{fig:results_ideal_points} shows the RMSE across the training ($N_{\mathrm{train}}=230$), validation ($N_{\mathrm{valid}}=179$), and test set ($N_{\mathrm{test}}=103$), and the first column block of Tab.~\ref{tab:errors_all_querystrain} lists the average over the respective set. The median values over all $n$ are $4.33\cdot 10^{-3}$ (training set), $4.30\cdot 10^{-3}$ (validation set), and $4.35\cdot 10^{-3}$ (test set). The relative variabilites across all $n$ are 2.74\% (training set), 2.76\% (validation set), and 2.39\% (test set), indicating little variability of the number of query points $n\in[5, 15000]$ on the errors.

\begin{table*}[!htp]
\begin{center}
\caption{Average root mean squared errors (RMSE) over the respective geometry cohort of idealized models. Comparison of influence of number of query points and strain augmentation of the loss function.}\label{tab:errors_all_querystrain}
\resizebox{1.0\columnwidth}{!}{%
\begin{tabular}{r|ccccccccccccc|cc}
$\varnothing\mathcal{J}_{u}$ & \multicolumn{9}{|l}{Idealized Geometries} \\\hline
 & \multicolumn{13}{|l}{\textit{Query Points}} & \multicolumn{2}{|l}{\textit{Strain}, $n=2500$} \\
 & \multicolumn{13}{|l}{$\mathbf{x},\boldsymbol{\upxi},\boldsymbol{\mu}_{g}(\{\ell,d,w\}),\lambda_{\mathrm{s}}=0$} & \multicolumn{2}{|l}{$\mathbf{x},\boldsymbol{\upxi},\boldsymbol{\mu}_{g}(\{\ell,d,w\})$} \\\hline
$ \cdot 10^{-2}$ & \rotatebox{90}{$n=5$} & \rotatebox{90}{$n=10$} & \rotatebox{90}{$n=25$}& \rotatebox{90}{$n=50$}& \rotatebox{90}{$n=100$} & \rotatebox{90}{$n=250$} & \rotatebox{90}{$n=500$} & \rotatebox{90}{$n=1000$} & \rotatebox{90}{$n=2500$} & \rotatebox{90}{$n=5000$} & \rotatebox{90}{$n=7500$} & \rotatebox{90}{$n=10000$} & \rotatebox{90}{$n=15000$} & \rotatebox{90}{$\lambda_{\mathrm{s}}=0$} & \rotatebox{90}{$\lambda_{\mathrm{s}}=0.1$} \\\hline
TRAIN & 0.680 & 0.558 & 0.445 & 0.432 & 0.420 & 0.427 & 0.433 & 0.441 & 0.428 & 0.437 & 0.403 & 0.441 & 0.425 & 0.440 & 0.399\\
VALID & 0.674 & 0.553 & 0.444 & 0.429 & 0.417 & 0.424 & 0.430 & 0.436 & 0.425 & 0.433 & 0.400 & 0.438 & 0.422 & 0.438 & 0.396\\
TEST & 0.668 & 0.552 & 0.446 & 0.435 & 0.422 & 0.430 & 0.435 & 0.442 & 0.429 & 0.437 & 0.406 & 0.441 & 0.426 & 0.443 & 0.401\\
\end{tabular}
}%
\end{center}
\end{table*}

\paragraph{Strain information}
The effects of enriching the loss function Eq.~(\ref{eq:Jw_u}) by strains directly computed from the finite element space are investigated. For this, the number of query points is set to $n=2500$, and the parameter $\lambda_{\mathrm{s}}$ in the combined loss Eq.~(\ref{eq:Jw}) is chosen such that displacement and strain losses are of comparable magnitude. The input layer of the network consists of $\mathbf{x},\boldsymbol{\upxi},\boldsymbol{\mu}_{g}(\{\ell,d,w\})$. Figure~\ref{fig:results_ideal_strain}a shows the \emph{displacement} loss on the training, validation, and test set for $\lambda_{\mathrm{s}}=0$ (bypassing strain computation) and $\lambda_{\mathrm{s}}=0.1$ (adding strain contributions to the loss), and Fig.~\ref{fig:results_ideal_strain}b shows the displacement RMSE over the epochs for both cases. The averages over the geometries are listed in the second column block of Tab.~\ref{tab:errors_all_querystrain}.

We observe mild improvements in the loss on the displacement prediction when adding strains to the overall loss function, amounting to a relative improvement of 9.32\% (training set), 9.59\% (validation set), and 9.48\% (test set). However, we observe an increase in computational expense by a factor of $\sim 5$ compared to bypassing gradient computations.

\subsubsection{Combined real and generated geometries}
\label{subsub:surrogate_real}
After studying different input layer settings on the idealized model cohort, we investigate the augmented dataset of real and generated geometries (44 + 976 models) and compare to training only on the real geometry cohort (44 models). For this study, we keep a fixed number of query points $n=15000$ and bypass strain contributions to the loss ($\lambda_{\mathrm{s}}=0$). We train our network using the following three variants of input layers:
\begin{itemize}
    \item $\mathbf{x},\boldsymbol{\upxi}$: Input layer consisting of Cartesian coordinates and UVCs;
        \item $\mathbf{x},\boldsymbol{\upxi},\boldsymbol{\mu}_{g}^{(\mathrm{D\text{-}SDF})}(\mathit{\Omega}_0)$: Input layer consisting of Cartesian coordinates, UVCs, and $\SDFmodel$ shape code of dimension 16;
    \item $\mathbf{x},\boldsymbol{\upxi},\boldsymbol{\mu}_{g}^{(\mathrm{PCA})}(\mathit{\Omega}_0)$: Input layer consisting of Cartesian coordinates, UVCs, and a $\PCAmodel$ shape code of dimension 34.
\end{itemize}

\begin{figure}[!htp]
\centering
\includegraphics[width=0.81\textwidth]{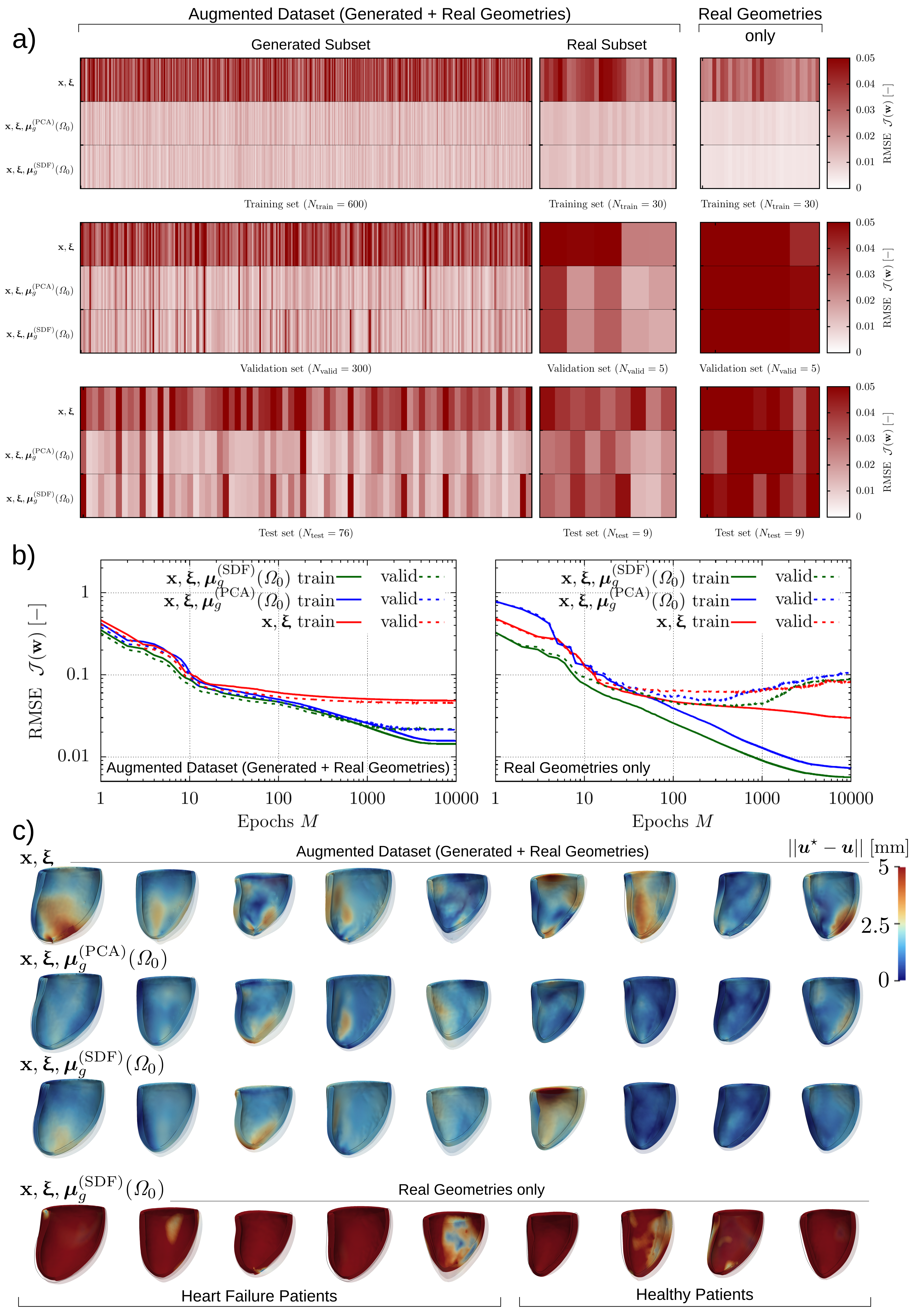}
\caption{Generated and real geometries: Augmented (real and generated geometries) vs. non-augmented (only real geometries) datasets. Comparison of different input layers and shape codes ($\mathbf{x},\boldsymbol{\upxi}$: only Cartesian coordinates plus UVCs; $\mathbf{x},\boldsymbol{\upxi},\boldsymbol{\mu}_{g}^{(\mathrm{PCA})}(\mathit{\Omega}_0)$: Cartesian coordinates, UVCs, $\PCAmodel$ shape code; $\mathbf{x},\boldsymbol{\upxi},\boldsymbol{\mu}_{g}^{(\mathrm{SDF})}(\mathit{\Omega}_0)$: Cartesian coordinates, UVCs, $\SDFmodel$ shape code). No strain contributions to the loss function ($\lambda_{\mathrm{s}}=0$), number of query points $n=15000$. a) RMSE on training, validation, and test set for three different input layers, split into generated and real geometries. b) RMSE over training epochs, split into training and validation loss. c) Absolute difference in ground-truth and inferred displacement on all nine test geometries from the real cohort.}\label{fig:results_genstrrod}
\end{figure}

Figure~\ref{fig:results_genstrrod}a shows the RMSE on the training ($N_{\mathrm{train}}=630$), validation ($N_{\mathrm{valid}}=305$), and test set ($N_{\mathrm{test}}=85$), Fig.~\ref{fig:results_genstrrod}b the RMSE over the number of training epochs (split into training and validation error), and Fig.~\ref{fig:results_genstrrod}c the plotted absolute difference in ground-truth and inferred displacement for all nine geometries from the test subset of the real cohort (HF and healthy). Results from the augmented (generated plus real) as well as the real cohort only are reported separately. The averages over the geometries are listed in the second (augmented) and third (real only) column block of Tab.~\ref{tab:errors_all_shapecode}. As for the idealized models, a substantial improvement is observable when adding shape codes retrieved from the $\PCAmodel$ and $\SDFmodel$ model compared to only considering $\mathbf{x}$ and $\boldsymbol{\upxi}$. Reductions in RMSE for the augmented dataset of generated and real geometries of 66.4\% and 68.9\% (training set), 58.5\% and 55.9\% (validation set), and 54.6\% and 40.5\% (test set) are achieved. Furthermore, we observe that training only on real geometries without dataset augmentation (last column block of Tab.~\ref{tab:errors_all_shapecode}) yields significantly worse results on the validation and test set, with a better training error, thus highlighting a severe overfitting when the proposed augmentation strategy is not employed.

\begin{table*}[!htp]
\begin{center}
\caption{Average root mean squared errors (RMSE) over the respective geometry cohort of idealized, generated, and real geometries. Comparison of influence of shape code and data augmentation.}\label{tab:errors_all_shapecode}
\begin{tabular}{r|ccccc|ccc|ccc}
$\varnothing\mathcal{J}_{u}$ & \multicolumn{5}{|l}{Idealized Geometries} & \multicolumn{3}{|l}{Gen. + Real Geometries} & \multicolumn{3}{|l}{Real Geometries} \\\hline
 & \multicolumn{11}{l}{\textit{Shape}} \\
 & \multicolumn{11}{l}{$n=15000,\lambda_{\mathrm{s}}=0$} \\\hline
$ \cdot 10^{-2}$ & \rotatebox{90}{$\mathbf{x}$} & \rotatebox{90}{$\mathbf{x},\boldsymbol{\upxi}$} & \rotatebox{90}{$\mathbf{x},\boldsymbol{\upxi},\boldsymbol{\mu}_{g}^{(\mathrm{PCA})}(\mathit{\Omega_{0}})$} & \rotatebox{90}{$\mathbf{x},\boldsymbol{\upxi},\boldsymbol{\mu}_{g}^{(\mathrm{SDF})}(\mathit{\Omega_{0}})$} & \rotatebox{90}{$\mathbf{x},\boldsymbol{\upxi},\boldsymbol{\mu}_{g}(\{\ell,d,w\})$} & \rotatebox{90}{$\mathbf{x},\boldsymbol{\upxi}$} & \rotatebox{90}{$\mathbf{x},\boldsymbol{\upxi},\boldsymbol{\mu}_{g}^{(\mathrm{PCA})}(\mathit{\Omega_{0}})$} & \rotatebox{90}{$\mathbf{x},\boldsymbol{\upxi},\boldsymbol{\mu}_{g}^{(\mathrm{SDF})}(\mathit{\Omega_{0}})$} & \rotatebox{90}{$\mathbf{x},\boldsymbol{\upxi}$} & \rotatebox{90}{$\mathbf{x},\boldsymbol{\upxi},\boldsymbol{\mu}_{g}^{(\mathrm{PCA})}(\mathit{\Omega_{0}})$} & \rotatebox{90}{$\mathbf{x},\boldsymbol{\upxi},\boldsymbol{\mu}_{g}^{(\mathrm{SDF})}(\mathit{\Omega_{0}})$} \\\hline
TRAIN & 4.772 & 2.856 & 0.637 & 0.609 & 0.425 & 3.926 & 1.318 & 1.221 & 2.914 & 0.722 & 0.561\\
VALID & 4.960 & 2.979 & 0.796 & 0.620 & 0.422 & 3.876 & 1.609 & 1.711 & 7.498 & 8.468 & 8.383 \\
TEST & 4.631 & 2.749 & 0.730 & 0.620 & 0.426 & 4.004 & 1.818 & 2.388 & 5.790 & 7.395 & 6.653 \\
\end{tabular}
\end{center}
\end{table*}


    \section{Discussion}
    \label{sec:discussion}
    
    In this work, we introduced a pipeline for the development of surrogate models for shape-dependent pa\-ra\-me\-trized PDEs, with specific application to cardiac mechanics. Because myocardial deformation depends nonlinearly on the domain geometry, clinically useful surrogate models must generalize to anatomies not observed during training. A central challenge in this context is the data-scarce regime typical of biomedical applications, where the cost and time required to acquire large collections of patient-specific geometries and simulations are often prohibitive. To mitigate data scarcity, we proposed decoupling geometry modeling from learning the physics response. A shape model is first trained on the available anatomies, and subsequently exploited in generative mode, to produce synthetic geometries representative of the distribution of the available shapes. These geometries are then used to augment the dataset of high-fidelity simulations employed to train the physics surrogate. Importantly, no additional real anatomies are required. The high-fidelity solver acts indeed as a bridge between synthetic geometry generation and physics learning, effectively introducing a stronger observational bias during surrogate model training. This strategy resulted in a marked improvement in generalization to unseen patients (error reduced from 6.7\% to 2.4\% for $\SDFmodel$). These results highlight that, in data-scarce settings, a well-trained shape model used in generative mode can effectively enrich the observed physical domain, and strengthen the surrogate model's generalization capability, without requiring any new real geometries.

A key structural feature of the proposed framework is the explicit separation between a shape model and a physics surrogate, conditioned on  a shape code. The shape model serves a dual role: (i) it learns the distribution of anatomies -- even in data-scarce regimes -- and enables sampling of novel shapes, and (ii) it provides a low-dimensional shape encoding that conditions the physics surrogate, thus allowing for generalization to unseen geometries without observable overfitting. For the surrogate model architecture, we adopted USM-Nets, a conditional neural-field supporting multiple positional encoding strategies. In this case, we used a domain-specific positional encoding, namely UVCs, which -- as shown by the results -- reduce the error of a na\"ive predictor receiving only spatial coordinates from 4.6\% to 2.7\% (in the idealized models), even without any shape encoding.
We compared two alternative shape encoding strategies: a PCA-based approach (generalizing statistical shape models to point-cloud representations) and a DeepSDF-based implicit neural representation trained in auto-decoder mode. 

Notably, the proposed decoupled framework enables a geometry-aware data augmentation strategy that is not naturally supported by several alternative approaches in the literature, such as GNN- or Transformer-based models that learn the mapping from geometry and physical parameters to the solution in a one-shot fashion. While such architectures have demonstrated strong approximation capabilities, their formulations typically rely on large training datasets to achieve stable generalization. For instance, recent graph-based and attention-based surrogate models report training regimes involving on the order of $10^3$--$10^4$ simulated samples to reach competitive accuracy \cite{taghizadeh2025multifidelity, sanchezgonzalez2020learning-physics-graph, serrano2024aroma}.
In contrast, our framework explicitly separates geometric representation from physics regression. This structural decoupling allows the shape model to learn an explicit latent distribution over anatomies, which can then be sampled to generate synthetic geometries for data augmentation. One-shot architectures that directly map geometry to solution do not explicitly model such a generative geometric distribution, and therefore do not naturally enable controlled sampling in the space of anatomies without additional modeling components.


The total error associated with our surrogate model can be decomposed into two components: the geometry encoding error and the physics prediction error. To disentangle these contributions, we introduced an idealized dataset in which an exact parameterization of the geometry is available. The results showed that the error introduced by shape encoding is, in this setting, smaller than the intrinsic physics prediction error. Specifically, normalized error increases from 0.43\% (exact parameterization) to 0.73\% (PCA-based encoding) and 0.62\% (DeepSDF-based encoding). The results also demonstrated robustness to point subsampling. Incorporating strain measurements alongside displacement in the loss function yielded a 9.5\% improvement in accuracy, at the expense of approximately fivefold increased computational cost, highlighting a trade-off between accuracy and efficiency.

Comparing the two shape encoding techniques (PCA-based and DeepSDF-based), we showed that both are robust to noise and data subsampling when reconstructing geometries via subsequent encoding and decoding. We introduced a quantity called the effective noise level, which captures how noise magnitude and number of points interact in determining shape model accuracy. In this framework, the PCA-based shape model yields slightly more accurate results when the effective noise level is very low, but is less resilient than the DeepSDF-based model as the effective noise level increases.

When integrated into the physics surrogate, the relative performance of the two encodings depended on the dataset. In the idealized setting, DeepSDF achieved lower error (0.62\% vs 0.73\%), whereas for real ventricular geometries PCA performed better (1.82\% vs 2.39\%). Overall, the two approaches yielded comparable performance for left ventricular geometries, indicating that both are adequate to capture the dominant modes of anatomical variability in this context. However, for more complex geometries whose variability cannot be well approximated by linear combinations of modes, implicit neural representations such as DeepSDF are expected to provide greater robustness. Additionally, PCA-based encoding requires point-to-point mesh correspondence -- enabled here through UVCs -- which may not be available in more general applications. From this perspective, DeepSDF-based encoding provides a more flexible alternative across different application scenarios compared to PCA-based encoding.

    \section{Methods}
    \label{sec:methods}

    \subsection{Data preprocessing}
\label{subsec:data_preprocessing}
\paragraph{Idealized data}
For benchmarking and testing purposes, a cohort of 512 idealized left ventricular geometries is created. Each ventricular geometry is modeled as a prolate ellipsoid, a shape commonly used to approximate the left ventricle due to its elongated geometry along the long axis. The ellipsoid is truncated along a plane orthogonal to the long axis, representing the basal opening toward the atrium. The determining geometric parameters are the ventricular long axis $\ell\in[75\;\mathrm{mm}, 115\;\mathrm{mm}]$, its maximum diameter $d\in[40\;\mathrm{mm}, 80\;\mathrm{mm}]$, and its transmural thickness $w\in[5\;\mathrm{mm}, 15\;\mathrm{mm}]$. From  each geometric parameter range, 8 evenly spaced samples are taken, each of which is combined with every other sample, yielding a $8 \times 8 \times 8$-sized cohort of geometries. Six exemplary models are shown in Fig.~\ref{fig:pipeline}b.

\paragraph{Patient-specific geometries}
In order to generate high-fidelity training data on patient-specific anatomical heart models, publicly available cohorts provided by Strocchi et al. \cite{strocchi2020} and Rodero et al. \cite{rodero2021} of volume-meshed full-heart cardiac geometries are used. Hearts in \cite{strocchi2020} are rather enlarged since they are from patients with diagnosed heart failure condition, whereas geometries in \cite{rodero2021} stem from ``healthy'' patients and do not exhibit pathological changes in shape. Here, we focus on inferring physics behavior on the left ventricle (LV), hence we take steps to extract, smooth, and re-mesh the LV subset for subsequent use in a finite element forward model.

Our geometry extraction and remeshing pipeline consists of the following steps:
\begin{itemize}
    \item Thresholding to extract all elements with LV region tag.
    \item Identification of 3 points on lower part of mitral valve rim for cut plane definition.
    \item Cut geometry at an inferior offset $\in [7.0\,\mathrm{mm}, 22.0\,\mathrm{mm}]$ to the defined plane (minimal distance to exclude all mitral valve rim artifacts) using a clipped threshold filter.
    \item Extract surface mesh on cut geometry, discarding volume mesh information (due to degenerate elements after cutting operation).
    \item Perfom 500 smoothing iterations on the surface mesh using the Smooth Polydata Filter.
    \item (Re-)Generation of volume tetrahedral finite element mesh based off the surface mesh using a mesh size factor of $1.0$, i.e. keeping characteristic element edge length from the original data.
    \item Redefinition of mesh tags, assigning surface element mesh tags to lumen, baseplane, and epicardium in order to facilitate boundary condition application in finite element model.
\end{itemize}
Six exemplary patient-specific geometries are shown in Fig.~\ref{fig:pipeline}a.


\subsection{Data alignment \& normalization}
\label{subsec:data_alignment}
To reduce the dimensionality of the anatomical space, we factor out rigid-body degrees of freedom and decouple size from the shape reconstruction problem, as size is cohort-dependent and would otherwise confound geometric comparisons. We note that the physical problems solved on the reconstructed geometries are translation- and rotation-invariant, but not scale-invariant; consequently, scaling information must be retained and incorporated into the surrogate model $\NNphysics$ to preserve correct physical behavior.

Each geometry is preprocessed as follows. First, it is centered at the LV barycenter and oriented so that the basal plane is horizontal, by performing PCA on the basal plane and retaining the first two principal components. After ensuring that the apex points downward and that the coordinates system is right-handed, we rotate the geometry around the z-axis in order to ensure that the right ventricle (RV) barycenter lays on the x-axis with positive x. Finally, the mesh is centered according to the barycenter of the LV and rescaled so that all vertices belong to the unit cube $[-1, 1]^3$. The scaling factor is stored as additional input for the shape reconstruction model $\NNsdf$ and for the surrogate model $\NNphysics$. The detailed procedure is reported in Algorithm~\ref{algo:mesh_preprocessing}.

\begin{algorithm}[H]
\SetAlgoLined
\caption{Orientation and Normalization of Mesh $\mathcal{M}$\label{algo:mesh_preprocessing}}
\KwData{Mesh $\mathcal{M}$ containing LV, RV, base plane segmentations}
\KwResult{Aligned and normalized mesh $\widetilde{\mathcal{M}}$}

\textbf{Step 1: Extract anatomical vertex sets}\;
\Indp
Basal plane $V_{\text{bas}}$, LV $V_{\text{lv}}$, RV $V_{\text{rv}}$, surface $V_{\text{surf}}$\;
\Indm

\textbf{Step 2: Initial centering}\;
\Indp
Compute the barycenter of LV $\mu_  {\text{lv}}$ and center all sets: $V \leftarrow V - \mu_{\text{lv}}$\;
\Indm

\textbf{Step 3: PCA alignment}\;
\Indp
Perform PCA on $V_{\text{bas}}$ to obtain principal components $P \in \mathbb{R}^{3 \times 3}$\;
Transform all vertex sets: $V' = VP^T$\;
Recenter on basal plane by subtracting the basal plane barycenter $ \mu_{\text{bas}}$: $V' \leftarrow V' - \mu_{\text{bas}}'$\;
\Indm

\textbf{Step 4: Orient apex-base axis}\;
\Indp
\If{$\text{mean}(V_{\text{lv}}'^{(:,2)}) > \text{mean}(V_{\text{bas}}'^{(:,2)})$}{
    Flip all coordinates: $V' \leftarrow -V'$ and $P \leftarrow -P$\;
}
Ensure right-handed system: \textbf{if} $\det(P) < 0$ \textbf{then} $V'^{(:,1)} \leftarrow -V'^{(:,1)}$\;
\Indm

\textbf{Step 5: Align RV to positive x-axis}\;
\Indp
Compute RV barycenter $\mu_{\text{rv}}'$ and rotation matrix $R_{\text{align}}$\;
Apply rotation: $V' \leftarrow V'R_{\text{align}}^T$ for all sets\;
\If{$\text{mean}(V_{\text{rv}}'^{(:,0)}) < 0$}{
    Apply 180° rotation around z-axis\;
}
\Indm

\textbf{Step 6: Final normalization}\;
\Indp
Center on LV by subtracting the new barycenter $\mu_{\text{lv}}'$: $V'' = V' - \mu_{\text{lv}}'$\;
Compute max coordinate: $s_{\max} = \max_{V,i,j} |V''_{ij}|$\;
Scale to unit cube: $\tilde{V} = V''/s_{\max}$\;
Store scaling factor $s_{\max}$\;
\Indm

\Return{Normalized and oriented mesh $\widetilde{\mathcal{M}}$ with scaling factor $s_{\max}$}
\end{algorithm}

\begin{figure}[htbp]
\centering
\begin{subfigure}{0.48\textwidth}
\centering
\includegraphics[width=\linewidth]{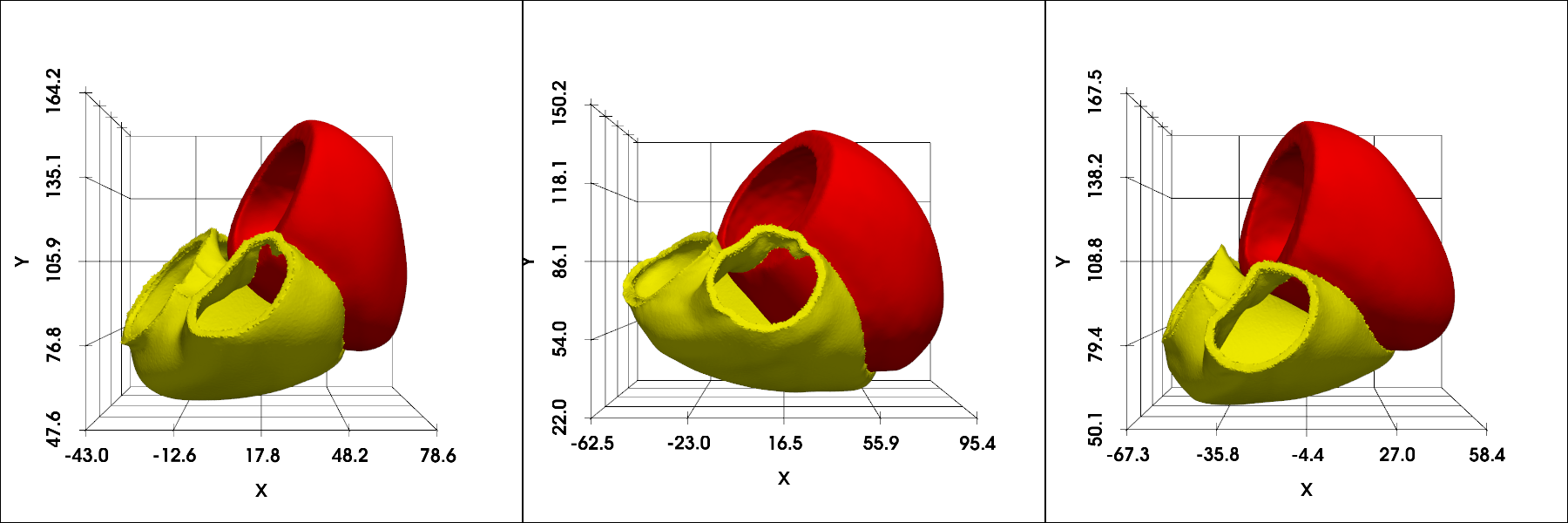} 
\caption{Heart failure cohort pre-alignment (XY-view)}
\label{fig:strocchi_pre_al}
\end{subfigure}
\hfill
\begin{subfigure}{0.48\textwidth}
\centering
\includegraphics[width=\linewidth]{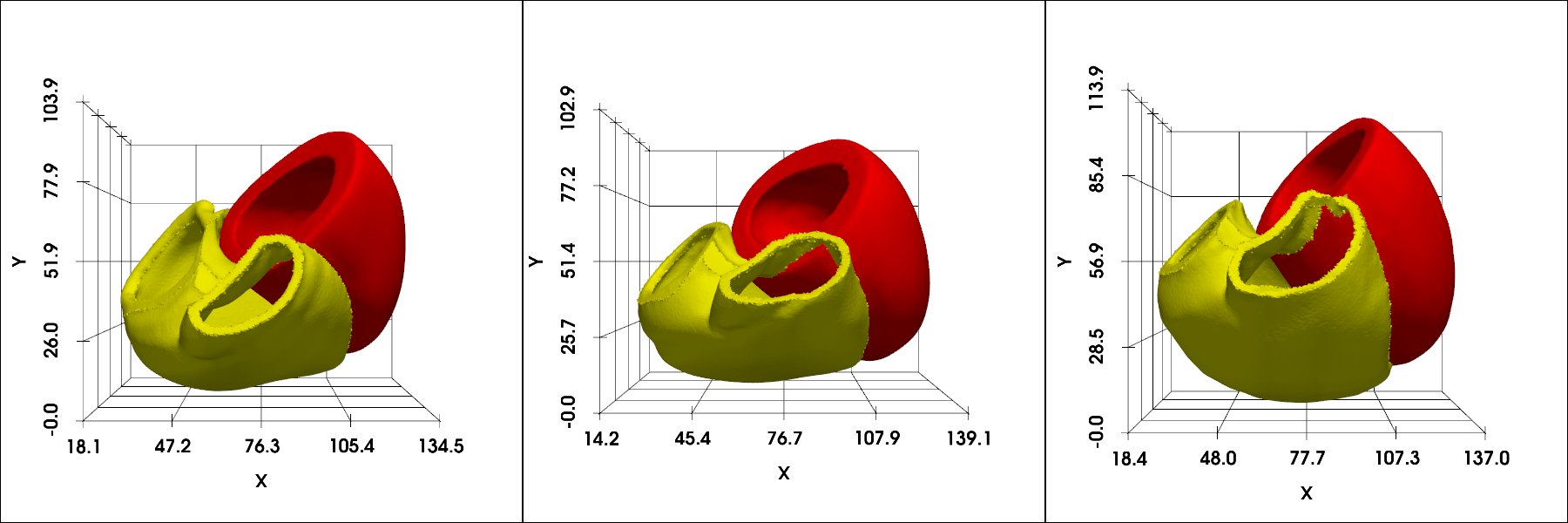}
\caption{Healthy cohort pre-alignment (XY-view)}
\label{fig:rodero_pre_al}
\end{subfigure}

\vspace{0.5cm} 

\begin{subfigure}{0.48\textwidth}
\centering
\includegraphics[width=\linewidth]{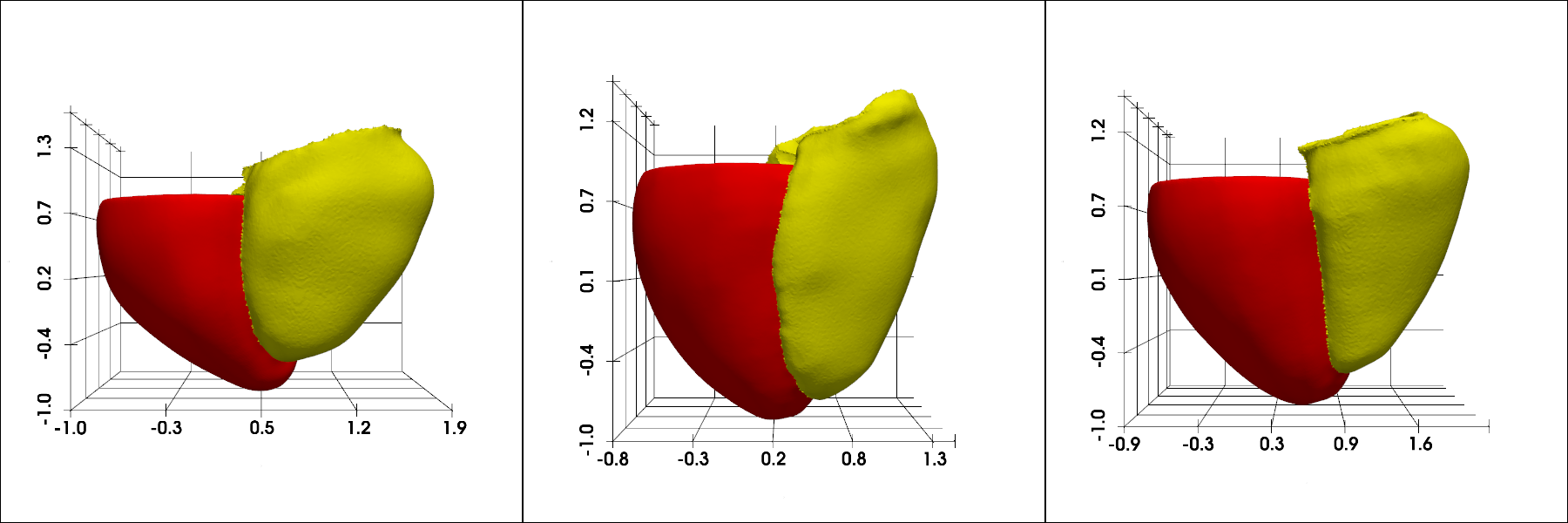}
\caption{Heart failure cohort post-alignment (XZ-view)}
\label{fig:strocchi_post_al}
\end{subfigure}
\hfill
\begin{subfigure}{0.48\textwidth}
\centering
\includegraphics[width=\linewidth]{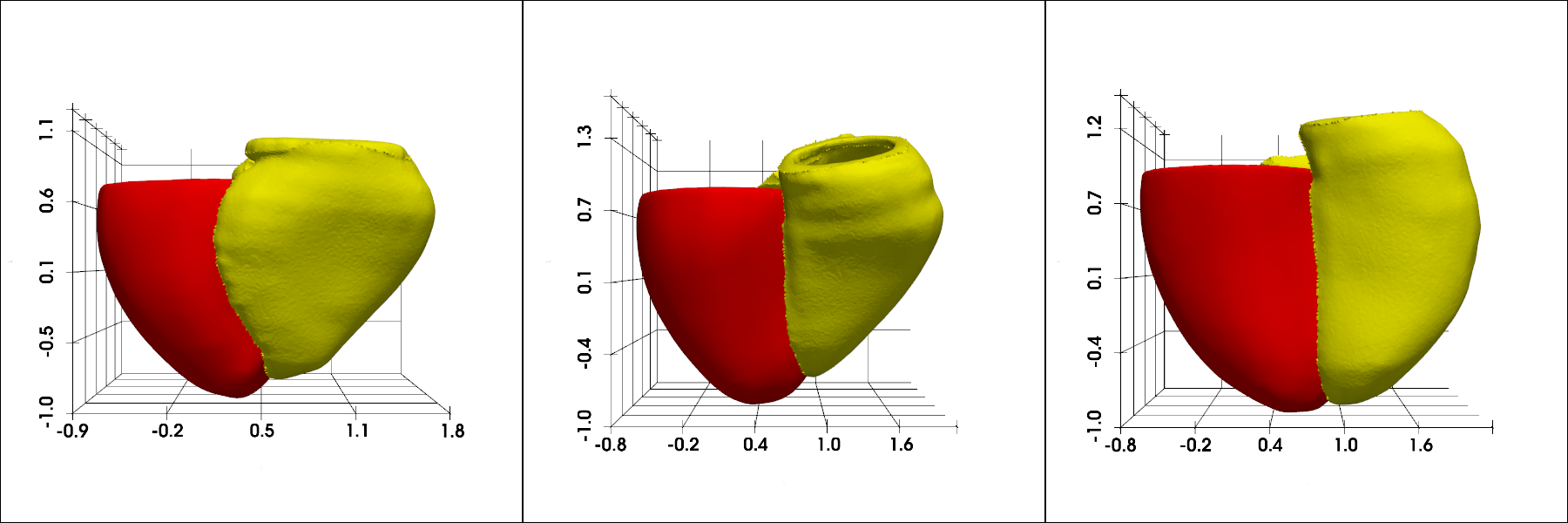}
\caption{Healthy cohort post-alignment (XZ-view)}
\label{fig:rodero_post_al}
\end{subfigure}

\caption{Views of anatomies from both cohorts pre and post alignment and orientation procedure.}
\label{fig:data_alignment_pre_post}
\end{figure}

\subsection{Dataset generation}
\label{subsec:dataset_generation}
The dataset consists of 44 left ventricular geometries divided into training (35 geometries: 19 from \cite{strocchi2020}, 16 from \cite{rodero2021}) and test (9 geometries: 5 from \cite{strocchi2020}, 4 from \cite{rodero2021}) sets. Additionally, 512 idealized prolate ellipsoid geometries are used, divided into training ($N_{\text{train}} = 409$) and test ($N_{\text{test}} = 103$) sets.

\paragraph{Point cloud sampling strategy.}
For training $\SDFmodel$, we construct a point cloud dataset $\mathcal{X}_i$ for each geometry following the sampling strategy from \cite{park2019deepsdf}. This strategy combines three types of samples to capture different aspects of the geometry:
\begin{itemize}
    \item 2000 points sampled on the surface and perturbed with Gaussian noise $\sigma = 0.025$
    \item 1500 points sampled on the surface and perturbed with Gaussian noise $\sigma = 0.25$
    \item 500 points sampled uniformly in the bounding box $[-1,1]^3$
\end{itemize}

A validation set is generated by perturbing 1000 points from the training surfaces with $\sigma = 0.25$.

For each sampled point, the ground-truth SDF value is computed as the distance to its closest neighbor on the original (unperturbed) surface, obtained via KD-tree queries \cite{maneewongvatana1999analysisapproximatenearestneighbor}. The sign is determined by checking whether the point is contained within a cell of the original mesh.

To preserve scale information factored out during preprocessing (Algorithm~\ref{algo:mesh_preprocessing}), each geometry $i$ is associated with a normalized scaling parameter $\mu_i$ used as additional input to condition both $\NNsdf$ and $\NNphysics$:
\begin{equation}
    \label{eq:normalized_max}
    \geoparam_i = \frac{s_{i_{\max}}}{\max_{j} s_{j_{\max}}},
\end{equation}
where $s_{i_{\max}}$ is the maximum absolute coordinate value of geometry $i$ before normalization.

\paragraph{Inference datasets.}
At inference time, latent codes are optimized using point clouds sampled directly on the geometry surfaces without perturbation. For real anatomies, 4000 surface points are used; for idealized geometries, 2000 surface points are used. Target SDF values are set to zero for all surface points.

\paragraph{Robustness evaluation dataset.}
To evaluate reconstruction quality under varying noise and sampling conditions (Section~\ref{subsubsec:quality_evaluation}), points are sampled on the surface and perturbed with Gaussian noise, but the ground-truth SDF value is retained as the noise-free value corresponding to the original surface point. This introduces a labeling error that simulates realistic measurement noise: the number of sampled points varies across $\{125, 250, 500, 1000, 2000\}$ and the noise standard deviation across $\{0.0, 0.0125, 0.025, 0.05, 0.1\}$.


\subsection{Shape model training step}
\label{subsec:shape_model_training}
\begin{figure}[!htp]
\centering
\includegraphics[width=0.8\textwidth]{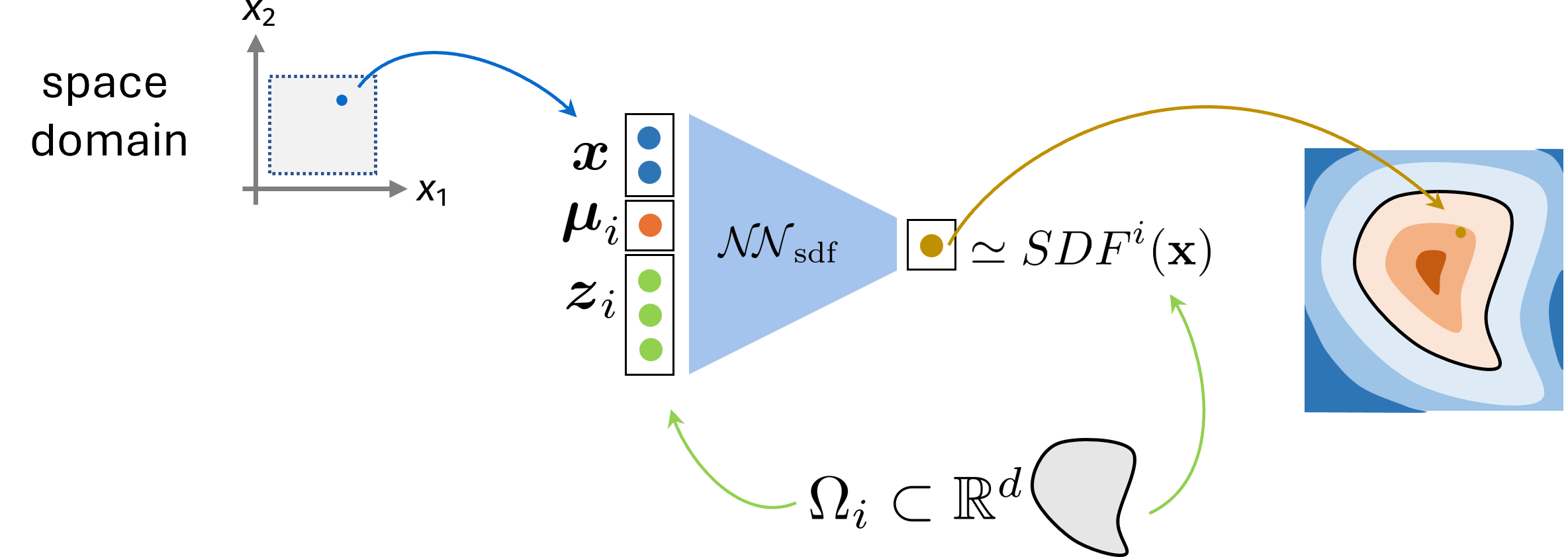}
\caption{Auto-decoder architecture visualization. The geometrical input $\point$ is concatenated with the latent code $\shapecode_i$ associated to the geometry $\Omega_i$ and with optional additional physical parameters $\mu_i$.}\label{fig:autodecoder}
\end{figure}

In this work, we adopt an autodecoder formulation to discover a continuous latent shape space jointly with a neural implicit SDF model. 
We choose to represent each 3D shape $\Omega_i$ implicitly by training the autodecoder to approximate its signed distance function (${\SDF}^i$):

\begin{equation}
\label{eq:sdf_def}
    \begin{aligned}
    {\SDF}^i(\point) = \begin{cases}
        - d(\point, \partial\Omega_i) \quad &\text{if  }  \point \in \Omega_i \\
        d(\point, \partial\Omega_i) \quad &\text{if  }  \point \notin \Omega_i, \\
    \end{cases} \\
    d(\point, \partial\Omega_i) := \inf_{\mathbf{y} \in \partial\Omega_i}  d(\point,\mathbf{y}) \quad \forall \point \in X.
    \end{aligned}
\end{equation}
This formulation has been shown to provide smooth latent spaces, high-fidelity reconstructions, and strong generalization across shape categories, as demonstrated by the DeepSDF method \cite{park2019deepsdf}.
The underlying mathematical formulation is based on a probabilistic perspective \cite{park2019deepsdf, liu2022learning}.

Given a set of $N$ shapes, each represented by the associated signed distance function $\{\SDF^i\}_{i=1}^N$, we build $N$ sets of $K_i$ point samples, each corresponding to one shape:

\begin{equation}
\label{eq:dataset_definition}
    X_i = \{(\point_j, \mu_i, s_j)_{j=1}^{K_i}: s_j = \SDF^i(\point_j)\}.
\end{equation}
In the auto-decoder setting, each geometry is paired with a shape code $\shapecode_i$ lying in the lower-dimensional latent space $\mathcal{Z}$.  The shape code $\shapecode_i$ is learned together with the parameters $\weightNNsdf$, which parametrize the SDF likelihood.

In the Bayesian setting, the posterior distribution over the latent variable $\shapecode_i$ can be decomposed as the product of a SDF likelihood term and a prior distribution $p(\shapecode_i)$ over the latent space, as in Eq.~\eqref{eq:posterior_decomposition}: 

\begin{equation}
    \label{eq:posterior_decomposition}
    p_{\weightNN}(\shapecode_i| X_i) \propto p(\shapecode_i)\Pi_{(\point_j, \mu_i, s_j) \in X_i} p_{\weightNN}(s_j|\shapecode_i; \point_j, \mu_i),
\end{equation}
where $\weightNN$ parametrizes the $\SDF$ likelihood. In our case, the $\SDF$ is approximated by using a neural network $\NNsdf$, described by its set of weights $\weightNNsdf$. As a consequence, the likelihood function can be defined as:

\begin{equation}
    \label{eq:likelihood_function}
    p_{\weightNNsdf}(s_j|\shapecode_i; \point_j, \mu_i) = \text{exp}(-\mathcal{L}(\NNsdf(\shapecode_i, \point_j, \mu_i; \weightNNsdf), s_j)),
\end{equation}
where $\mathcal{L}$ is a custom loss function.
The prior distribution over the latent space is modeled as a multivariate Gaussian with zero-mean and spherical covariance $\sigma_{\shapecode}^{2} I$ providing a simple and well-behaved initial structure for the latent manifold.

At training time, the goal is to find both the best neural network approximation of the SDF associated to each shape and the best corresponding shape code in the latent space. This can be obtained by minimizing the joint log-posterior for all shapes with respect both to the codes $\{\shapecode_i\}_{i=1}^{N}$ and the network weights $\weightNNsdf$ such that:

\begin{equation}
    \label{eq:joint_log-posterior}
    \min_{\weightNNsdf, \{\shapecode_i\}_{i=1}^{N}} \sum_{i=1}^{N}\left(\sum_{j=1}^{K} \mathcal{L}(\NNsdf(\shapecode_i, \point_j, \mu_i; \weightNNsdf), s_j) + \frac{1}{2\sigma_{\shapecode}^2} \|\shapecode_i\|_{2}^{2}\right).
\end{equation}
Here, $\mathcal{L}(\Tilde{s}_j, s_j)$ penalizes discrepancies between the predicted $\SDF$ value $\Tilde{s}_j = \NNsdf(\shapecode_i, \point_j; \weightNNsdf)$ and the ground-truth $\SDF$ value $s_j$. In this work we adopt an $\ell_1$-based loss, defined as:

\begin{equation}
    \label{eq:L1_loss}
    \mathcal{L}(\Tilde{s_j}, s_j) = \frac{|\Tilde{s_j}-s_j|}{b},
\end{equation}
where $b$ is a normalization constant.

To ensure smooth interpolation between shapes and stable latent space optimization, we control the Lipschitz constant on the decoder. This regularization bounds the sensitivity of $\SDF$ predictions to perturbations in the latent code, preventing discontinuous transitions in the manifold of learned shape codes.
To achieve this we apply a weight regularization technique based on minimizing the Lipschitz constant of the network introduced in \cite{liu2022learning}, and named Lipschitz Multilayer Perceptron. 

Applying the proposed architecture, each layer $l$ of $\NNsdf$ is augmented with a Lipschitz weight normalization layer, where the bound $c_l$ is computed from the current weights as:

\begin{equation}
    \label{ep:lipschitz_normalization:layer}
    y = \sigma(\hat{{\weightNNsdf}}_l\point+b_l), \quad \hat{{\weightNNsdf}}_l = \text{normalize}_{\infty}({\weightNNsdf}_l, \text{softplus}(c_l)), \quad  \text{softplus}(c_l) = \text{ln}(1+e^{c_l}),
\end{equation}
where ${\weightNNsdf}_l$ is the weight matrix for layer $l$, and $c_l = \|{\weightNNsdf}_l\|_{\infty}$ denotes the maximum absolute row sum of the weight matrix. The normalization operator $\text{normalize}_{\infty}$ scales each row of ${\weightNNsdf}_l$ such that its $\ell_1$ norm does not exceed $\text{softplus}(c_l)$, ensuring the layer's operator norm is bounded.

The set $\{c_l\}_{l=1}^{N_{\text{layers}}}$ is used to compute an upper bound for the Lipschitz constant of $\NNsdf$ as:

\begin{equation}
\label{eq:overall_lipschitz_constant}
C(\weightNNsdf) = \prod_{l=1}^{N_{\text{layers}}}\text{softplus}(c_l).    
\end{equation}
Given the Lipschitz constant of the network $C(\weightNNsdf)$ the final training objective function becomes:

\begin{equation}
    \label{eq:lipschitz_loss_functional_minimization}
    \min_{\weightNNsdf, \{\shapecode_i\}_{i=1}^{N}} \sum_{i=1}^{N}\left(\avsum_{j=1}^K{|\NNsdf(\shapecode_i, \point_j, \mu_i; \weightNNsdf)-s_j|} + w_{\text{prior}} \|\shapecode_i\|_{2}^{2} + w_{\text{lip}} C(\weightNNsdf) \right),
\end{equation}
where $w_{\text{lip}}$ is a hyperparameter controlling the strength of the Lipschitz regularization and $\avsum_{j=1}^K$ is the averaged sum over the set of indices $\{1, \dots, j, \dots, K\}$.

\subsection{Shape model inference step}
\label{subsec:shape_model_inference}
At inference time, the decoder weights $\weightNNsdf$ are fixed, and we want to find the optimal latent code $\shapecode_i$ for a new shape. While in training we used a zero-mean spherical Gaussian prior, it is now possible to incorporate the empirical distribution of learned latent codes to obtain a more informative prior.

From a Bayesian perspective, observing that the training codes $\{\shapecode_i\}_{i=1}^{N}$ form an empirical distribution with mean $\boldsymbol{\mu}$ and covariance $\Sigma$, the natural choice for the prior at inference is $p(\shapecode) = \mathcal{N}(\boldsymbol{\mu}, \Sigma)$. Substituting this into the posterior distribution (analogous to~\eqref{eq:posterior_decomposition}) and taking the negative log-posterior leads to the Mahalanobis regularization term:

\begin{equation}
    \label{eq:mahalanobis_prior_term}
    \|\shapecode_i\|_{M}^2
    =
    (\shapecode_i - \boldsymbol{\mu})^{\top} \Sigma^{-1} (\shapecode_i - \boldsymbol{\mu}).
\end{equation}

The Maximum a Posteriori (MAP) estimate of the latent code for a single geometry $X_i = \{(\point_j, \mu_i, s_j)\}_{j=1}^{K}$ is therefore obtained by solving
\begin{equation}
\label{eq:map_inference_mahalanobis}
\shapecode_i^{\star}
=
\argmin_{\shapecode_i}
\left[
    \avsum_{j=1}^{K}{\left| \NNsdf(\shapecode_i, \point_j, \mu_i; \weightNNsdf) - s_j \right|}
    +
    w_{\mathrm{post}}\;(\shapecode_i - \boldsymbol{\mu})^{\top}\Sigma^{-1}(\shapecode_i - \boldsymbol{\mu})
\right],
\end{equation}
where $w_{\mathrm{post}} = \frac{b}{2K}$ balances the reconstruction term against fidelity to the learned latent distribution. The factor $\frac{1}{2}$ comes from the Gaussian log-likelihood, while $b$ is the normalization constant from~\eqref{eq:L1_loss}, and $K$ accounts for the averaging over sample points.

We observe that $w_{\mathrm{post}} = \frac{b}{2K}$ decreases with the number of sample points $K$ and increases with the noise scaling factor $b$. As a consequence, with scarce or noisy observations, the optimization relies more heavily on the learned prior distribution $\mathcal{N}(\boldsymbol{\mu}, \Sigma)$, regularizing the latent code toward the training manifold. On the other hand, with abundant high-quality data, the reconstruction term dominates, allowing the model to accurately fit the observed geometry.

\subsection{Shape reconstruction metrics}
\label{subsec:shape_reconstruction_metrics}
In order to evaluate the quality of the reconstructed shapes we consider 5 metrics. 

\paragraph{Chamfer Distance ($\CD$).}
The Chamfer distance measures the dissimilarity between two point clouds $\mathcal{D}_1$ and $\mathcal{D}_2$ by computing the average nearest-neighbor distance in both directions:

\begin{equation}
\label{eq:chamfer_distance}
\CD(\mathcal{D}_1, \mathcal{D}_2) = \frac{1}{2} \left(\frac{1}{|\mathcal{D}_1|}\sum_{\mathbf{x} \in \mathcal{D}_1} \min_{\mathbf{y} \in \mathcal{D}_2} \|\mathbf{x} - \mathbf{y}\|_2 + \frac{1}{|\mathcal{D}_2|}\sum_{\mathbf{y} \in \mathcal{D}_2} \min_{\mathbf{x} \in \mathcal{D}_1} \|\mathbf{y} - \mathbf{x}\|_2\right).
\end{equation}

\paragraph{Normalized Chamfer Distance ($\CDnorm$).}
Since the model is trained on shapes that are normalized in the cube $[-1, 1]^3$, in order to factor out cohort size differences, we also compute the Chamfer distance between reconstruction and normalized target. This provides a scale-invariant measure of shape similarity.

\paragraph{Mean Level Estimate ($\MLE$).}
To assess how closely the network predicts zero on the actual surface, we compute the Mean Level Estimate as the average absolute SDF prediction across all ground-truth surface vertices and all geometries:
\begin{equation}
\label{eq:mle}
    \MLE = \avsum_{i=1}^{N_{\text{geo}}} \left(\avsum_{\mathbf{x} \in \mathbf{X}_i} |\NNsdf(\shapecode_i, \mathbf{x}, \mu_i; \weightNNsdf)|\right),
\end{equation}
where $\mathbf{X}_i$ denotes the set of surface vertices for the $i$-th geometry. Since ground-truth surface points should have SDF value zero, the MLE indicates the bias of the learned SDF representation across the dataset. Lower MLE values indicate better alignment between the predicted zero-level sets and the true geometry surfaces.

\paragraph{Intersection over Union ($\IoU$).}
We evaluate the decoder $\NNsdf$ on a fine regular grid and classify each grid point as inside (SDF $< 0$) or outside (SDF $\geq 0$) the reconstructed geometry. Similarly, we determine the inside/outside classification for each grid point with respect to the target mesh. The IoU is then defined as:

\begin{equation}
\label{eq:iou}
\IoU = \frac{|V_{\text{pred}} \cap V_{\text{target}}|}{|V_{\text{pred}} \cup V_{\text{target}}|},
\end{equation}
where $V_{\text{pred}}$ and $V_{\text{target}}$ denote the sets of grid points classified as inside the predicted and target geometries, respectively.

\paragraph{Dice Coefficient ($\Dice$).}
The Dice coefficient, also known as the F1 score or Sørensen–Dice coefficient, provides an alternative volumetric overlap measure:

\begin{equation}
\label{eq:dice}
\Dice = \frac{2|V_{\text{pred}} \cap V_{\text{target}}|}{|V_{\text{pred}}| + |V_{\text{target}}|}.
\end{equation}
The Dice coefficient is more sensitive to changes when the overlap is high, making it particularly useful for assessing the reconstruction quality in medical imaging applications, where accurate volume estimation is critical.
We use a 3D grid with $96\times96\times96$ resolution for computation of $\IoU$ and $\Dice$.

\subsection{Shape generation}
\label{subsec:shape_generation}
Once the decoder $\NNsdf$ and latent codes $\{\shapecode_i\}_{i=1}^{N}$ have been optimized, we can extract explicit mesh representations of the learned shapes.

For each training shape $i$, we evaluate the conditioned network $\NNsdf(\shapecode_i, \cdot; \weightNNsdf)$ on a dense 3D grid in the domain of interest. The resulting discrete SDF values are then processed using the marching cubes algorithm \cite{lorensen1987marching} to extract a triangular mesh corresponding to the zero-level set, which represents the shape boundary $\partial\Omega_i$.

We can exploit the learned latent space to generate new plausible geometries. We sample latent codes from the empirical distribution $\mathcal{N}(\boldsymbol{\mu}, \Sigma)$ of the training codes, where $\boldsymbol{\mu}$ and $\Sigma$ are the mean and covariance calculated from $\{\shapecode_i\}_{i=1}^{N}$. Each sampled code $\shapecode_{\text{new}} \sim \mathcal{N}(\boldsymbol{\mu}, \Sigma)$ is used to condition the decoder, which is then evaluated on the same 3D grid and processed via marching cubes to obtain a mesh.
We also sample a scaling factor $\mu_{\text{new}}$ from a uniform distribution obtained by taking the minimum and maximum scaling factors used during training.

Under the assumption that the latent space is sufficiently smooth and $\NNsdf$ exhibits regularity with respect to the latent code (enforced by the Lipschitz regularization in~\eqref{eq:lipschitz_loss_functional_minimization}), the sampled codes produce realistic novel geometries that interpolate between training examples. These synthetic shapes can be used for data augmentation in downstream tasks, such as surrogate modeling of ventricles.

\subsection{Architecture/Training/Specifics}
\label{subsec:specifics}
\paragraph{Model architecture.}
The decoder $\NNsdf$ is a fully connected network with 6 hidden layers of 64 neurons each. We use the Exponential Linear Unit (ELU) activation function instead of ReLU to ensure $C^1$ continuity of the learned SDF. This smoothness is necessary for downstream finite element simulations, as ReLU activations introduce sharp angles in the reconstructed zero-level set that can cause numerical instabilities.

Following~\cite{rodero2021,unberath2015open}, we set the latent space dimension to $\dim(\mathcal{Z}) = 16$, when working with real patient anatomies. 

Idealized geometries are generated by varying a set of 3 parameters: the ventricular long axis $l$, the maximum diameter $d$ and transmural thickness $w$. 
When training the reconstruction model on them we choose to keep the same architectural structure used for the real setting and thus, knowing that the geometrical scaling value $\mu_i$ correlates strongly with $l$, we set $\dim(\mathcal{Z}) = 2$. In this way the scaling factor acts as third dimension of the latent space.
Latent codes are initialized by sampling from $\mathcal{N}(\mathbf{0}, 10^{-2}\mathbf{I})$.

\paragraph{Hyperparameters.}
Following~\eqref{eq:lipschitz_loss_functional_minimization}, we set the prior weight $w_{\text{prior}} = 10^{-4}$ and the Lipschitz weight $w_{\text{lip}} = 10^{-6}$ when training the model on patient anatomies and $w_{\text{lip}} = 10^{-9}$, when training on idealized geometries. The prior weight encourages latent codes to remain within a bounded region, while the Lipschitz regularization promotes smoothness in the latent space for stable interpolation.
The posterior weight for the inference loss is set to $w_{\text{post}}=6.25 \e{-5}$.

\paragraph{Training and inference.}
The $\NNsdf$ model is trained for up to 12000 epochs with ADAM optimizer and decaying learning rate starting from $5 \e{-3}$. Training is automatically stopped when the root mean squared error computed on the validation points stagnates for more than 200 epochs.

At inference time the weights of the network $\weightNNsdf$ are frozen and only the latent code associated with the geometry of interest is trained via the procedure indicated in section~\ref{subsec:shape_model_inference}. We train each inference code for 2000 epochs with ADAM optimizer. 

\subsection{PCA encoding fitting} 
\label{subsec:pca_encoding}
\paragraph{1-to-1 correspondence via universal ventricular coordinates} Dimensionality reduction via principal component analysis (PCA) requires establishing a standardized coordinate system where 1-to-1 correspondence between different patient anatomies can be achieved. We employ UVCs \cite{bayer2018universal}, a cylindrical-like parameterization $(\zeta, \varrho, \phi)$, where $\zeta \in [0,1]$ represents the apicobasal direction, $\varrho \in [0,1]$ denotes the transmural direction, and $\phi \in [-\pi, \pi]$ corresponds to the rotational coordinate. In this space we define a uniform sampling grid with dimension $N_{\text{features}}=N_\zeta \times N_\varrho \times N_\phi$, creating a regular lattice of query points.

The reparametrization works as follows.The mesh vertices of each mesh, represented in Cartesian coordinates $(x,y,z)$, are associated with their corresponding UVCs. Spatial interpolation is performed to map the irregular patient-specific mesh onto the uniform UVC grid. Using a nearest-neighbour interpolation method, we obtain three interpolated fields $\hat{x}(\zeta, \varrho, \phi)$, $\hat{y}(\zeta, \varrho, \phi)$, and $\hat{z}(\zeta, \varrho, \phi)$ each evaluated at the grid points. At this point, we filter the grid to retain only surface points, corresponding to $z=1$ (base) or $\varrho \in \{0, 1\}$ (endocardium and epicardium). The coordinates are concatenated and flattened to obtain a final vector representation of length $N_{\text{features}}$.
The vector representations of all train geometries are concatenated to form the snapshot matrix $\mathbf{S}_{\text{train}}$ with shape $3N_{\text{features}} \times N_{\text{train}}$.

\paragraph{PCA via singular value decomposition}

The snapshot matrix is first centered by subtracting the mean feature vector over all train patients:
\begin{equation}
\tilde{\mathbf{S}}_{\text{train}} = \mathbf{S}_{\text{train}} - \bar{\mathbf{s}}\mathbf{1}^T \quad \bar{\mathbf{s}} = \frac{1}{M}\sum_{i=1}^{N_{\text{train}}} \mathbf{s}_i.
\end{equation}
Then, the centered snapshot matrix is decomposed via truncated SVD:
\begin{equation}
\tilde{\mathbf{S}}_{\text{train}} = \mathbf{U}\mathbf{\Sigma}\mathbf{V}^T,
\end{equation}
where $\mathbf{U} \in \mathbb{R}^{3N_{\text{features}} \times r}$ contains the left singular vectors (spatial modes), $\mathbf{\Sigma} \in \mathbb{R}^{r \times r}$ is the diagonal matrix of singular values $\sigma_i$, $\mathbf{V}^T \in \mathbb{R}^{r \times N_{\text{train}}}$ contains the right singular vectors, and $r = \min(3N_{\text{features}}, N_{\text{train}})$ is the rank of the decomposition.

The cumulative explained variance ratio is computed as:
\begin{equation}
\eta_k = \frac{\sum_{i=1}^{k} \sigma_i^2}{\sum_{i=1}^{r} \sigma_i^2}.
\end{equation}
This metric quantifies the proportion of geometric variability captured by the first $k$ principal modes. In our work we select $k=r-1$ in order to retain as much geometrical information as possible, since even small variations could be relevant for downstream tasks. The final mode is excluded to avoid numerical artifacts.

The reduced basis $\mathbf{U}_k$ is constructed by selecting the first $k$ left singular vectors.
Each patient geometry is encoded into a low-dimensional latent representation via projection onto the principal component basis:
\begin{equation}
\mathbf{z}_i = \mathbf{U}_k^T \tilde{\mathbf{s}}_i \in \mathbb{R}^k,
\end{equation}
where $\tilde{\mathbf{s}}_i$ is the centered feature vector.
The original geometry can be approximately reconstructed as:
\begin{equation}
\hat{\mathbf{s}}_i = \bar{\mathbf{s}} + \mathbf{U}_k\mathbf{c}_i,
\end{equation}
where $\mathbf{c}_i$ is the shape code for patient $i$. $\hat{\mathbf{s}}_i$ is then reshaped in the original tridimensional space.

\subsection{PCA encoding inference}
\label{subsec:pca_encoding_infernce}
Since UVCs cannot be precomputed on point cloud data, shape codes need to be inferred through an optimization procedure.
Given a set of observed surface points $\mathbf{X} \in \mathbb{R}^{N_{\text{obs}} \times 3}$ sampled from the test geometry, we look for the optimal shape code $\mathbf{z}^* \in \mathbb{R}^k$ that best reconstructs the observed data, by exploiting the PCA basis introduced in Section~\ref{subsec:pca_encoding}.
The reconstruction model is defined as:
\begin{equation}
\hat{\mathbf{s}}(\shapecode) = \bar{\mathbf{s}} + \mathbf{U}_k \shapecode.
\end{equation}
The flattened reconstruction $\hat{\mathbf{s}}(\shapecode)$ is reshaped to obtain the surface point cloud $\hat{\mathbf{s}}_{\text{surf}}(\shapecode) \in \mathbb{R}^{N_{\text{features}} \times 3}$.

The objective function for the optimization procedure is the $\CD$ between the surface points and the geometry reconstructed by the PCA model:

\begin{equation}
\shapecode^* = \argmin_{\shapecode} \CD(\mathbf{X}, \hat{\mathbf{s}}_{\text{surf}}(\shapecode)).
\end{equation}
The optimization is performed using the Adam optimizer with learning rate $\alpha=5\e{-2}$ for 2000 epochs. At each step of the optimization, gradients are computed via automatic differentiation.

\subsection{Forward PDE problem}
\label{subsec:forward_pde_problem}
\paragraph{Finite strain mechanics} We endeavor to infer the deformation due to diastolic filling of the LV at defined cavity pressure $\hat{p}$. Therefore, ground truth data is generated by solving the following quasi-static finite strain solid mechanics boundary value problem on each of the (idealized and patient-specific) cardiac geometries. The strong form of the boundary value problem defined on an LV geometry $\mathit{\Omega}_0$, cf. Fig.~\ref{fig:pipeline}e for an exemplary patient-specific model, reads as follows: 

\begin{equation}
\label{eq:lv_bvp}
\begin{aligned}
\boldsymbol{0} &= \nabla_{0} \cdot (\boldsymbol{FS}) &&\text{in} \; \mathit{\Omega}_{0}, \\
(\boldsymbol{FS})\boldsymbol{n}_{0} &= -\hat{p}\,J\boldsymbol{F}^{-\mathrm{T}}\boldsymbol{n}_{0} &&\text{on} \; \mathit{\Gamma}_{0}^{N}, \\
(\boldsymbol{FS})\boldsymbol{n}_{0} &= k_{\mathrm{b}}^{||}\,(\boldsymbol{I}-\boldsymbol{n}_{0}\otimes\boldsymbol{n}_{0})\boldsymbol{u} \\
&  + k_{\mathrm{b}}^{\perp}(\boldsymbol{n}_{0}\otimes\boldsymbol{n}_{0}) \boldsymbol{u} &&\text{on} \; \mathit{\Gamma}_{0}^{R,\mathrm{b}}, \\
(\boldsymbol{FS})\boldsymbol{n}_{0} &= k_{\mathrm{e}}^{||}\,(\boldsymbol{I}-\boldsymbol{n}_{0}\otimes\boldsymbol{n}_{0})\boldsymbol{u} \\
&  + k_{\mathrm{e}}^{\perp}(\boldsymbol{n}_{0}\otimes\boldsymbol{n}_{0}) \boldsymbol{u} &&\text{on} \; \mathit{\Gamma}_{0}^{R,\mathrm{e}},
\end{aligned}
\end{equation}
where $\boldsymbol{F}=\boldsymbol{I} + \nabla_0\boldsymbol{u}$ is the deformation gradient, $J=\det\boldsymbol{F}$ its determinant, and $\boldsymbol{u}$ the unknown displacement field from the (stress-free) reference configuration $\mathit{\Omega}_0$ to a deformed configuration $\mathit{\Omega}$. Further, $\mathit{\Gamma}_{0}^{N}$ and $\mathit{\Gamma}_{0}^{R,(\cdot)}$ are Neumann and Robin boundaries, respectively, and $\boldsymbol{n}_0$ is the outward normal with respect to the reference frame. The constitutive relation for the 2nd Piola-Kirchhoff stress tensor $\boldsymbol{S}$ is of anisotropic hyperelastic type:
\begin{equation}
\label{eq:pk2}
\begin{aligned}
\boldsymbol{S} = 2\frac{\partial\mathit{\Psi}(\boldsymbol{C})}{\partial\boldsymbol{C}},
\end{aligned}
\end{equation}
where $\boldsymbol{C}=\boldsymbol{F}^{\mathrm{T}}\boldsymbol{F}$ is the right Cauchy-Green deformation tensor and $\mathit{\Psi}$ the strain energy density function, here given by the anisotropic Holzapfel-Ogden model \cite{holzapfel2009}:
\begin{equation}
\label{eq:holz}
\begin{aligned}
\mathit{\Psi} &= \frac{a_0}{2b_0}\left(e^{b_0(\bar{I}_C - 3)} - 1\right) + \sum\limits_{i\in\{f,s\}}\frac{a_i}{2b_i}\left(e^{b_i(I_{4,i}-1)^2}-1\right) + \frac{a_{fs}}{2b_{fs}}\left(e^{b_{fs}I_{8}^2} - 1\right) + \frac{\kappa}{2}(J-1)^{2},
\end{aligned}
\end{equation}
where $\bar{I}_C = J^{-2/3}\mathrm{tr}\boldsymbol{C}$ is the isochoric first invariant of the Cauchy-Green tensor, and $I_{4,f} = \boldsymbol{f}_0 \cdot \boldsymbol{C}\boldsymbol{f}_0$ and $I_{4,s} = \boldsymbol{s}_0 \cdot \boldsymbol{C}\boldsymbol{s}_0$ are invariants defining the squared stretches in fiber and sheet directions $\boldsymbol{f}_0$ and $\boldsymbol{s}_0$, respectively, while $I_8 = \boldsymbol{f}_0 \cdot \boldsymbol{C}\boldsymbol{s}_0$ describes the shear stretch between the two (initially perpendicular) directions. Fiber and sheet directions are generated using rule-based approaches \cite{bayer2012,doste2019}. Figure~\ref{fig:pipeline}e shows the transmurally varying fiber field $\boldsymbol{f}_0$ along with its angle $\alpha$ with respect to the ventricle's circumference. Finally, $\kappa$ represents a bulk modulus, chosen such that deformation is nearly incompressible.\\

We solve problem Eq.~(\ref{eq:lv_bvp}) using the finite element method, seeking displacements $\boldsymbol{u}\in \boldsymbol{\mathcal{V}}^{D,h}$ such that
\begin{equation}
\label{eq:lv_bvp_weak}
\begin{aligned}
&\int\limits_{\mathit{\Omega}_0}\boldsymbol{S}(\boldsymbol{C}(\boldsymbol{u})):\frac{1}{2}\delta\boldsymbol{C}(\delta\boldsymbol{u})\,\mathrm{d}V + \int\limits_{\mathit{\Gamma}_{0}^{N}}\hat{p}\,J(\boldsymbol{u})\boldsymbol{F}(\boldsymbol{u})^{-\mathrm{T}}\boldsymbol{n}_{0}\cdot\delta\boldsymbol{u}\,\mathrm{d}A + \\
&\int\limits_{\mathit{\Gamma}_{0}^{R,\mathrm{b}}}[k_{\mathrm{b}}^{||}\,(\boldsymbol{I}-\boldsymbol{n}_{0}\otimes\boldsymbol{n}_{0})\boldsymbol{u} + k_{\mathrm{b}}^{\perp}(\boldsymbol{n}_{0}\otimes\boldsymbol{n}_{0}) \boldsymbol{u}]\cdot\delta\boldsymbol{u}\,\mathrm{d}A +
\int\limits_{\mathit{\Gamma}_{0}^{R,\mathrm{e}}}[k_{\mathrm{e}}^{||}\,(\boldsymbol{I}-\boldsymbol{n}_{0}\otimes\boldsymbol{n}_{0})\boldsymbol{u} + k_{\mathrm{e}}^{\perp}(\boldsymbol{n}_{0}\otimes\boldsymbol{n}_{0}) \boldsymbol{u}]\cdot\delta\boldsymbol{u}\,\mathrm{d}A = 0
\end{aligned}
\end{equation}
for all displacement test functions $\delta\boldsymbol{u} \in \boldsymbol{\mathcal{V}}^{h}$, where $\boldsymbol{\mathcal{V}}^{D,h}$ and $\boldsymbol{\mathcal{V}}^{h}$ are suitable trial and test spaces (identical up to potential Dirichlet conditions included in the trial space). Throughout this study, first-order piecewise polynomial spaces on tetrahedral meshes are used. Parameters of the constitutive model Eq.~(\ref{eq:holz}), and the boundary conditions in Eq.~(\ref{eq:lv_bvp}), are given in Tab.~\ref{tab:parameters_bvp}.

\begin{table*}[!htp]
\begin{center}
\caption{Parameters of the constitutive model Eq.~(\ref{eq:holz}) as well as boundary conditions in Eq.~(\ref{eq:lv_bvp}).}\label{tab:parameters_bvp}
\begin{tabular}{ccccccccc}
$a_{0}\;[\mathrm{kPa}]$ & $b_{0}\;[-]$ & $a_{f}\;[\mathrm{kPa}]$ & $b_{f}\;[-]$ & $a_{s}\;[\mathrm{kPa}]$ & $b_{s}\;[-]$ & $a_{fs}\;[\mathrm{kPa}]$& $b_{fs}\;[-]$ & $\kappa\;[\mathrm{kPa}]$ \\\hline
$0.059$ & $8.023$ & $18.472$ & $16.026$ & $2.481$ & $11.120$ & $0.216$ & $11.436$ & $10^{2}$ \\\hline\hline
$k_{\mathrm{b}}^{\perp}\;[\mathrm{kPa}]$ & $k_{\mathrm{b}}^{||}\;[\mathrm{kPa}]$ & $k_{\mathrm{e}}^{\perp}\;[\mathrm{kPa}]$ & $k_{\mathrm{e}}^{||}\;[\mathrm{kPa}]$ & $\hat{p}\;[\mathrm{kPa}]$ & & & & \\\hline
$2\cdot 10^{3}$ & $10^{2}$ & $10$ & $1$ & $2.67$ & & & & 
\end{tabular}
\end{center}
\end{table*}

\paragraph{Universal ventricular coordinates (UVC)} We compute the continuous fields of UVCs -- the apicobasal, transmural, and circumferential coordinates $\zeta$, $\varrho$, and $\phi$, respectively -- by solving three Laplace problems \cite{bayer2018universal} on $\mathit{\Omega}_{0}$ (cf. Fig.~\ref{fig:pipeline}e for boundary labels). For the apicobasal coordinate, we solve:
\begin{equation}
\label{eq:laplace_zeta}
\begin{aligned}
-\nabla \cdot (\nabla \zeta) &= 0  &&\text{in} \; \mathit{\Omega}_{0}, \\
\zeta &=0  &&\text{on} \; P_{\mathrm{apex}}, \\
\zeta &=1  &&\text{on} \; \mathit{\Gamma}^{R,\mathrm{b}}, 
\end{aligned}
\end{equation}
where $P_{\mathrm{apex}}$ is the epicardial node of the left ventricular apex. Note that here, we do not perform geodesic distance sampling of $\zeta$, in contrast to \cite{bayer2018universal}.

The transmural coordinate $\varrho$ varies between 0 and 1 (between endo- and epicardium), hence it is a solution to:
\begin{equation}
\label{eq:laplace_rho}
\begin{aligned}
-\nabla \cdot (\nabla \varrho) &= 0  &&\text{in} \; \mathit{\Omega}_{0}, \\
\varrho &=1  &&\text{on} \; \mathit{\Gamma}_{0}^{R,\mathrm{e}}, \\
\varrho &=0  &&\text{on} \; \mathit{\Gamma}_{0}^{N}.
\end{aligned}
\end{equation}

Finally, the circumferential coordinate $\phi$ varies between $-\pi$ and $\pi$ and is solution of the following Laplace problem:
\begin{equation}
\label{eq:laplace_phi}
\begin{aligned}
-\nabla \cdot (\nabla \phi) &= 0  &&\text{in} \; \mathit{\Omega}_{0}, \\
\phi &=\pi  &&\text{on} \; \mathit{\Gamma}_{0}^{c}, \\
\phi &=-\pi &&\text{on} \; \mathit{\Gamma}_{0}^{c\prime}.
\end{aligned}
\end{equation}
Therein, $\mathit{\Gamma}_{0}^{c}$ is a coronary cut plane spanned by the apex, the center of the base place, and the apex point projected onto the base plane, and $\mathit{\Gamma}_{0}^{c\prime}$ is a plane rotated about an infinitesimal angle $\mathrm{d}\phi$.\\
Problems Eq.~(\ref{eq:laplace_zeta})--(\ref{eq:laplace_phi}) are discretized and solved using piecewise linear finite elements. The UVC fields are depicted in Fig.~\ref{fig:pipeline}h.

\subsection{Surrogate model}
\label{subsec:surrogate_model}

In order to predict the deformation field of ventricles under diastolic pressure loading conditions, a fully connected neural network (FCNN), or multi-layer perceptron (MLP), for learning the discrete field of ventricular displacements on the nodal points of a LV geometry is introduced. The FCNN is trained on the cohorts of (idealized or patient-specific) cardiac geometries, and its input layer consists of (min-max normalized versions) of the following quantities:
\begin{itemize}
    \item $\mathbf{x}\in\mathbb{R}^{n\times 3}$: the discrete vector of spatial cartesian coordinates;
    \item $\boldsymbol{\upxi}\in\mathbb{R}^{n\times 3}$: the discrete vector of UVCs;
    \item $\boldsymbol{\mu}_{g}(\mathit{\Omega}_0)\in\mathbb{R}^{m}$: a shape code encoding geometric variability (e.g., $\PCAmodel$, $\SDFmodel$);
    \item $\boldsymbol{\mu}_{p}\in\mathbb{R}^{n_{\mathrm{p}}}$: a set representative of the parameterization of the physics model (here invariant).
\end{itemize}

The output layer of the FCNN is the discrete vector of displacements $\mathbf{u}$.
The architecture of the network is depicted Fig.~\ref{fig:pipeline}j.
The discrete set of cartesian coordinates $\mathbf{x}$, UVCs $\boldsymbol{\upxi}$, and displacements $\mathbf{u}$ are defined according to:
\begin{align}
	\mathbf{x}_{i}=\begin{bmatrix} \boldsymbol{x}_{1}\\\vdots\\\boldsymbol{x}_{n} \end{bmatrix}_{i} = \begin{bmatrix} x_{1} \; y_{1} \; z_{1} \\ \vdots \\ x_{n} \; y_{n} \; z_{n} \end{bmatrix}_{i}, \qquad
    \boldsymbol{\upxi}_{i}=\begin{bmatrix} \boldsymbol{\xi}_{1}\\\vdots\\\boldsymbol{\xi}_{n} \end{bmatrix}_{i} = \begin{bmatrix} \zeta_{1} \; \varrho_{1} \; \sin\phi_{1} \; \cos\phi_{1} \\ \vdots \\ \zeta_{n} \; \varrho_{n} \; \sin\phi_{n} \; \cos\phi_{n} \end{bmatrix}_{i}, \qquad
    \mathbf{u}_{i}=\begin{bmatrix} \boldsymbol{u}_{1}\\\vdots\\\boldsymbol{u}_{n} \end{bmatrix}_{i} = \begin{bmatrix} u_{x,1} \; u_{y,1} \; u_{z,1} \\ \vdots \\ u_{x,n} \; u_{y,n} \; u_{z,n} \end{bmatrix}_{i},
\end{align}
where the index $n$ corresponds to the number of discrete points (nodes) of the $i$-th geometry. Further, $x,y,z$ denote the axes of the cartesian reference frame, and $\zeta\in[0,1]$, $\rho\in[0,1]$, and $\phi\in[-\pi,\pi]$ are the universal apicobasal, transmural, and circumferential ventricular coordinates. Due to the sharp discontinuity of $\phi$ from $-\pi$ to $\pi$, we prefer to use continuous representations $\sin\phi$ and $\cos\phi$, instead.

For scaling and performance purposes, we perform a min-max normalization of the input data as follows. Minimum and maximum values of geometry and displacement are computed as follows:
\begin{align}
	& x_{\min} = \min\left\{\min\begin{bmatrix} ||\boldsymbol{x}_{1}||\\\vdots\\ ||\boldsymbol{x}_{n}|| \end{bmatrix}_{1}, \hdots, \min\begin{bmatrix} ||\boldsymbol{x}_{1}||\\\vdots\\ ||\boldsymbol{x}_{n}|| \end{bmatrix}_{N}\right\}, \quad
	x_{\max} = \max\left\{\max\begin{bmatrix} ||\boldsymbol{x}_{1}||\\\vdots\\ ||\boldsymbol{x}_{n}|| \end{bmatrix}_{1}, \hdots, \max\begin{bmatrix} ||\boldsymbol{x}_{1}||\\\vdots\\ ||\boldsymbol{x}_{n}|| \end{bmatrix}_{N}\right\},
\end{align}
\begin{align}
	& u_{\min} = \min\left\{\min\begin{bmatrix} ||\boldsymbol{u}_{1}||\\\vdots\\ ||\boldsymbol{u}_{n}|| \end{bmatrix}_{1}, \hdots, \min\begin{bmatrix} ||\boldsymbol{u}_{1}||\\\vdots\\ ||\boldsymbol{u}_{n}|| \end{bmatrix}_{N}\right\}, \quad
	u_{\max} = \max\left\{\max\begin{bmatrix} ||\boldsymbol{u}_{1}||\\\vdots\\ ||\boldsymbol{u}_{n}|| \end{bmatrix}_{1}, \hdots, \max\begin{bmatrix} ||\boldsymbol{u}_{1}||\\\vdots\\ ||\boldsymbol{u}_{n}|| \end{bmatrix}_{N}\right\},
\end{align}
and the min-max normalized data for the $i$-th observation is
\begin{align}
	\widehat{\mathbf{x}}_{i}=\frac{1}{x_{\max}-x_{\min}}\begin{bmatrix} \boldsymbol{x}_{1} - x_{\min}\mathbf{1}\\\vdots\\\boldsymbol{x}_{n} - x_{\min}\mathbf{1} \end{bmatrix}_{i} = \begin{bmatrix} \widehat{\boldsymbol{x}}_{1}\\\vdots\\\widehat{\boldsymbol{x}}_{n} \end{bmatrix}_{i}, \qquad \widehat{\mathbf{u}}_{i}=\frac{1}{u_{\max}-u_{\min}}\begin{bmatrix} \boldsymbol{u}_{1} - u_{\min}\mathbf{1}\\\vdots\\\boldsymbol{u}_{n} - u_{\min}\mathbf{1} \end{bmatrix}_{i} = \begin{bmatrix} \widehat{\boldsymbol{u}}_{1}\\\vdots\\\widehat{\boldsymbol{u}}_{n} \end{bmatrix}_{i},
\end{align}
so that values range $\in [0,1]$.
The UVC fields are min-max normalized per component\footnote{Note that $\zeta,\varrho\in[0,1]$ already, hence these components are not affected by min-max normalization.} as follows: 
\begin{align}
	\widehat{\boldsymbol{\upxi}}_{i}=\frac{\boldsymbol{\upxi}_{i} - \boldsymbol{\upxi}_{\min}}{\boldsymbol{\upxi}_{\max}-\boldsymbol{\upxi}_{\min}},
\end{align}
with the vector of minimum and maximum UVCs being
\begin{align}
    \boldsymbol{\upxi}_{\min} &= \left[\min\left\{(\boldsymbol{\upxi}_{1})_{1},\hdots,(\boldsymbol{\upxi}_{N})_{1}\right\}, \hdots, \min\left\{(\boldsymbol{\upxi}_{1})_{4},\hdots,(\boldsymbol{\upxi}_{N})_{4}\right\}\right]^{\mathrm{T}},\\
    \boldsymbol{\upxi}_{\max} &= \left[\max\left\{(\boldsymbol{\upxi}_{1})_{1},\hdots,(\boldsymbol{\upxi}_{N})_{1}\right\}, \hdots, \max\left\{(\boldsymbol{\upxi}_{1})_{4},\hdots,(\boldsymbol{\upxi}_{N})_{4}\right\}\right]^{\mathrm{T}}.
\end{align}
Analogously, the min-max normalized vectors of shape codes for the $i$-th geometry and parameters for the $k$-th parametric configuration read
\begin{align}
	\widehat{\boldsymbol{\mu}}_{g,i}=\frac{\boldsymbol{\mu}_{g,i} - \mu_{g,\min}\mathbf{1}}{\mu_{g,\max}-\mu_{g,\min}}, \qquad \widehat{\boldsymbol{\mu}}_{p,k}=\frac{\boldsymbol{\mu}_{p,k} - \boldsymbol{\mu}_{p,\min}}{\boldsymbol{\mu}_{p,\max}-\boldsymbol{\mu}_{p,\min}},
\end{align}
with the minimum and maximum shape code across geometries being $\mu_{g,\min} = \min\left\{\min\boldsymbol{\mu}_{g,1},\hdots,\min\boldsymbol{\mu}_{g,N}\right\}$ and $\mu_{g,\max} = \max\left\{\max\boldsymbol{\mu}_{g,1},\hdots,\max\boldsymbol{\mu}_{g,N}\right\}$. The vectors of minimum and maximum parameters read
\begin{align}
    \boldsymbol{\mu}_{p,\min} &= \left[\min\left\{(\boldsymbol{\mu}_{p,1})_{1},\hdots,(\boldsymbol{\mu}_{p,N_{\mathrm{p}}})_{1}\right\}, \hdots, \min\left\{(\boldsymbol{\mu}_{p,1})_{n_{\mathrm{p}}},\hdots,(\boldsymbol{\mu}_{p,N_{\mathrm{p}}})_{n_{\mathrm{p}}}\right\}\right]^{\mathrm{T}}, \\
    \boldsymbol{\mu}_{p,\max} &= \left[\max\left\{(\boldsymbol{\mu}_{p,1})_{1},\hdots,(\boldsymbol{\mu}_{p,N_{\mathrm{p}}})_{1}\right\}, \hdots, \max\left\{(\boldsymbol{\mu}_{p,1})_{n_{\mathrm{p}}},\hdots,(\boldsymbol{\mu}_{p,N_{\mathrm{p}}})_{n_{\mathrm{p}}}\right\}\right]^{\mathrm{T}}, 
\end{align}
with the number of parameters $n_{\mathrm{p}}$ and the number of parametric configurations $N_{\mathrm{p}}$ (here, we only consider $N_{\mathrm{p}}=1$).

The FCNN outputs the (min-max normalized) displacement $\widehat{\mathbf{u}}^{\star}$ at the set of query points $\widehat{\mathbf{x}}$,
\begin{align}
	\mathbf{u}^{\star}(\widehat{\mathbf{x}};\widehat{\boldsymbol{\mu}}_{p},\mathit{\Omega}_0) \simeq \NNphysics(\widehat{\mathbf{x}},\widehat{\boldsymbol{\upxi}},\widehat{\boldsymbol{\mu}}_{g}(\mathit{\Omega}_0), \widehat{\boldsymbol{\mu}}_{p}; \weightNNphysics),
\end{align}
with the trainable parameters (weights and biases) $\mathbf{w}$. We seek an optimal set of trainable parameters
\begin{align}
	\boldsymbol{\mathrm{w}}^{\star} = \argmin_{\boldsymbol{\mathrm{w}}}\mathcal{J}(\boldsymbol{\mathrm{w}}) 
\end{align}
being the minimizer of the loss function
\begin{align}
\mathcal{J}(\boldsymbol{\mathrm{w}}) = \mathcal{J}_{u}(\boldsymbol{\mathrm{w}}) + \lambda_{\mathrm{s}}\,\mathcal{J}_{\nabla u}(\boldsymbol{\mathrm{w}}),\label{eq:Jw}
\end{align}
which here is decomposed into the mean squared error loss relating to the displacement,
\begin{align}
	\mathcal{J}_{u}(\boldsymbol{\mathrm{w}}) = \frac{1}{N} \avsum\limits_{i=1}^{N} \left(g_{i}\frac{1}{n} \sum\limits_{j=1}^{n}||\widehat{\boldsymbol{u}}_{j}^{\star}-\widehat{\boldsymbol{u}}_{j}||^2 \right),\label{eq:Jw_u}
\end{align}
and one that penalizes displacement gradients, i.e. strains:
\begin{align}
	\mathcal{J}_{\nabla u}(\mathbf{w}) = \frac{1}{N} \avsum\limits_{i=1}^{N} \left(g_{i}\frac{1}{n_{\mathrm{e}}} \sum\limits_{j=1}^{n_{\mathrm{e}}}||\boldsymbol{E}(\boldsymbol{u}_{j}^{\star})-\boldsymbol{E}(\boldsymbol{u}_{j})||^2 \right),\label{eq:Jw_gradu}
\end{align}
where $g_i$ are weighting factors. Here, $\boldsymbol{E}$ is the vector of (elemental) Green-Lagrange strains in Voigt notation:
\begin{align}
\boldsymbol{E} = \begin{bmatrix}
	E_{11} & E_{22} & E_{33} & 2 E_{12} & 2 E_{13} & 2 E_{23}
\end{bmatrix}^{\mathrm{T}},
\end{align}
$n_{\mathrm{e}}$ is the number of cells per model $i$, and $N$ is the total number of observations, i.e. geometries. The regularization parameter $\lambda_{\mathrm{s}}$ in Eq.~(\ref{eq:Jw}) controls the contribution of gradient information that adds to the loss. Further, note that the strain is computed with the de-normalized displacement.

The displacement gradient here is computed using the linear tetrahedral finite element space, with the four nodal coordinates and displacements for one tetrahedron being
\begin{align}
\boldsymbol{x}^{(e)} = \begin{bmatrix}
	x^{1} & y^{1} & z^{1} \\
	x^{2} & y^{2} & z^{2} \\
	x^{3} & y^{3} & z^{3} \\
	x^{4} & y^{4} & z^{4}
\end{bmatrix} \qquad \text{and} \qquad \boldsymbol{u}^{(e)} = \begin{bmatrix}
	u_x^{1} & u_y^{1} & u_z^{1} \\
	u_x^{2} & u_y^{2} & u_z^{2} \\
	u_x^{3} & u_y^{3} & u_z^{3} \\
	u_x^{4} & u_y^{4} & u_z^{4}
\end{bmatrix},
\end{align}
respectively. The vector of nodal finite element shape functions and its derivatives in finite element reference space read
\begin{align}
\boldsymbol{N} = \begin{bmatrix}
	N^{1} \\ N^{2} \\ N^{3} \\ N^{4} 
\end{bmatrix} = \begin{bmatrix}
	1-\xi_{1}-\xi_{2}-\xi_{3} \\ \xi_{1} \\ \xi_{2} \\ \xi_{3}
\end{bmatrix} \qquad \text{and} \qquad  \boldsymbol{N}_{,\boldsymbol{\xi}} = \begin{bmatrix} -1 & -1 & -1 \\ 1 & 0 & 0 \\ 0 & 1 & 0 \\ 0 & 0 & 1\end{bmatrix},
\end{align}
yielding the element-centered displacement gradient
\begin{align}
\nabla_{0}\boldsymbol{u}^{(e)} = \boldsymbol{u}^{(e)^{\mathrm{T}}} \boldsymbol{N}_{,\boldsymbol{x}} = \boldsymbol{u}^{(e)^{\mathrm{T}}} \boldsymbol{N}_{,\boldsymbol{\xi}} \, \boldsymbol{J}^{-1} = \boldsymbol{u}^{(e)^{\mathrm{T}}} \boldsymbol{N}_{,\boldsymbol{\xi}} \, \left(\boldsymbol{x}^{(e)^{\mathrm{T}}}\boldsymbol{N}_{,\boldsymbol{\xi}}\right)^{-1}.
\end{align}
Finally, the element Green-Lagrange strain is 
\begin{align}
\boldsymbol{E}^{(e)} = \frac{1}{2}\left(\boldsymbol{F}^{(e)^\mathrm{T}}\boldsymbol{F}^{(e)} - \boldsymbol{I}\right),
\end{align}
with the deformation gradient $\boldsymbol{F}^{(e)} = \boldsymbol{I} + \nabla_{0}\boldsymbol{u}^{(e)}$.

\subsection{Surrogate training}
\label{subsec:surrogate_training}
Shape-informed inference of the displacement field from end-diastolic LV inflation is performed on two different geometry cohorts. We aim at quantifying the different types of shape encoding on the quality of the inferred displacement field, test the influence on the number of query points, and assess the impact of enriching the loss function by gradient information (strains).

First, we perform tests on the idealized cohort (Sec.~\ref{subsec:data_preprocessing}) of 512 geometries, cf. Fig.~\ref{fig:pipeline}b. The set is once randomly split into 230 training, 179 validation, and 103 test geometries. The same split is kept throughout all the studies on that cohort. Second, we investigate the performance of shape encoding on the augmented cohort of 44 real geometries (24 heart failure, 20 healthy), Fig.~\ref{fig:pipeline}c, and 976 generated models (Sec.~\ref{subsec:shape_generation}), cf. Fig.~\ref{fig:pipeline}d.\footnote{Out of 1000 generated heart models, 24 disqualified for meshing and model generation due to degenerate geometric features. See Appendix Fig.~\ref{fig:broken5} for some exemplary geometries which had to be discarded.} In order to balance loss contributions from real vs. generated models, we weigh the loss term of real geometries by $g_i=3$ compared to generated ones ($g_{i}=1$) in Eq.~(\ref{eq:Jw_u}). We compare this augmented dataset to using the real-only dataset of 44 models and show the benefits of data augmentation. Patient-specific and generated geometries are randomly split into 600 and 30 training, 300 and 5 validation, as well as 76 and 9 test geometries, respectively. Again, the same split is used on all studies with these cohorts.

Hyperparameter tuning is initially carried out for the idealized models and for the augmented generated plus real cohort individually. We assume a fixed number of three layers between input and output layer. We perform 500 epochs and evaluate the error on the validation set, with 5 startup trials and 50 warm-up steps, and seek the number of neurons $n_{\ell_1}\in[10,50]$, $n_{\ell_2}\in[5,40]$, and $n_{\ell_3}\in[3,20]$ in layer 1, 2, and 3, respectively. An optimal neural network architecture for the idealized cohort of $n_{\ell_1}=50$, $n_{\ell_2}=39$, $n_{\ell_3}=14$ is identified. For the augmented real and generated geometries, we identify the optimal numbers of neurons as $n_{\ell_1}=50$, $n_{\ell_2}=29$, and $n_{\ell_3}=20$.


We use hyperbolic tangent ($\tanh$) activation functions for each of the neurons, and training is performed using a BFGS optimizer with a maximum number of $M=10000$ epochs. For memory and efficiency reasons, we randomly subsample $n=15000$ spatial points (nodes) per geometry (if not indicated otherwise) and use the same subset of points for repeated, differently configured runs to achieve optimal comparability. 

\subsection{Software}
\label{subsec:software}
Data preprocessing for the shape modeling step was performed using \textit{SciPy} \cite{virtanen2020scipy} and \textit{scikit-learn} \cite{pedregosa2011scikit}. Geometry creation of the idealized models is done in \textit{FreeCAD} \cite{freecad2012}. Geometry reconstructions are carried out using \textit{vtk} \cite{schroeder2006vtk}. All meshing is performed in \textit{gmsh} \cite{geuzaine2009gmsh}. The finite element solution of problem Eq.~(\ref{eq:lv_bvp_weak}) is solved using \texttt{life\textsuperscript{x}} \cite{africa2022-lifex}. The shape model $\NNsdf$ and the mechanics surrogate model $\NNphysics$ were implemented in \textit{JAX} \cite{bradbury2018jax} and \textit{Flax} \cite{heek2020flax}, and trained using \textit{Optax} \cite{deepmind2020jax}. Hyperparameter tuning for $\NNphysics$ is carried out using \textit{Optuna} \cite{akiba2019-optuna}. Post-processing and mesh extraction were performed using \textit{scikit-image} \cite{vanderwalt2014skimage} and \textit{PyVista} \cite{sullivan2019pyvista}.


    \section*{Acknowledgements}

    The present research has been supported by the project PRIN2022, MUR, Italy, 2023-2025, P2022N5ZNP “SIDDMs: shape-informed data-driven models for parametrized PDEs, with application to computational cardiology”, funded by the European Union (Next Generation EU, Mission 4 Component 2).
    D.C., M.H., F.B., S.Pagani, and F.R. acknowledge the grant Dipartimento di Eccellenza 2023-2027, funded by MUR, Italy.
    D.C., F.B., S.Pagani, S.Pezzuto, and F.R. are members of GNCS, ``Gruppo Nazionale per il Calcolo Scientifico'' (National Group for Scientific Computing) of INdAM (Istituto Nazionale di Alta Matematica).
    F.B. acknowledges the “INdAM - GNCS Project”, codice CUP E53C24001950001.
    
	\begin{appendices}
		\section{Additional Materials}
        \label{sec:additional_materials}
        \begin{figure}[!htp]
\centering
\includegraphics[width=0.8\textwidth]{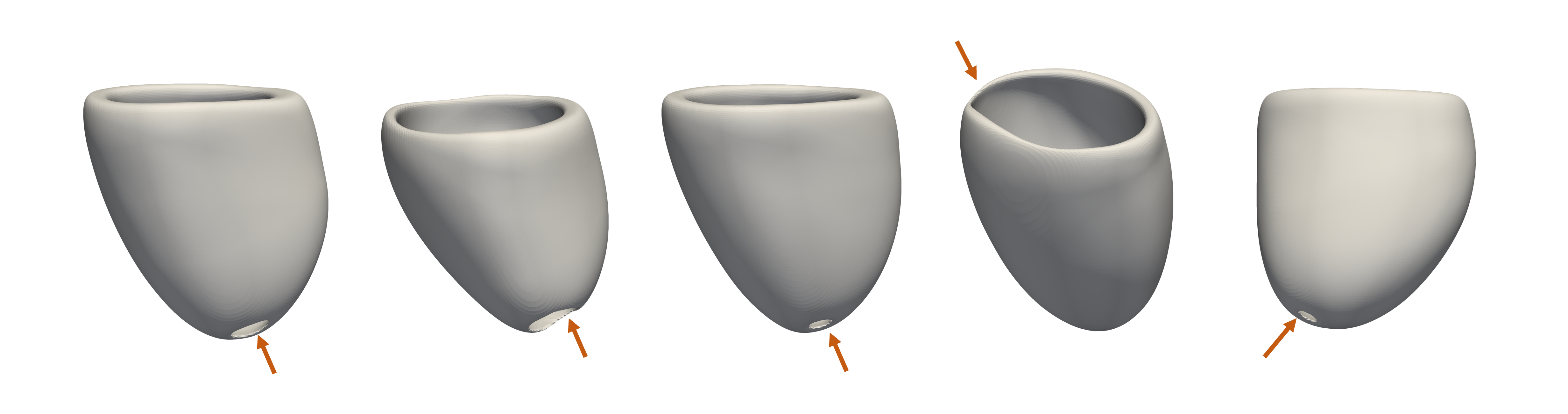}
\caption{Examples of five discarded synthetic geometries (samples 6, 39, 184, 280, 361) exhibiting artifacts near the apex or base.}
\label{fig:broken5} 
\end{figure}

Figure~\ref{fig:broken5} shows 5 examples of the discarded synthetic geometries. Out of 1000, 24 geometries were discarded mostly due to small holes close to the apex or degeneration at the cut plane level.

\begin{figure}[!htp]
\centering
\includegraphics{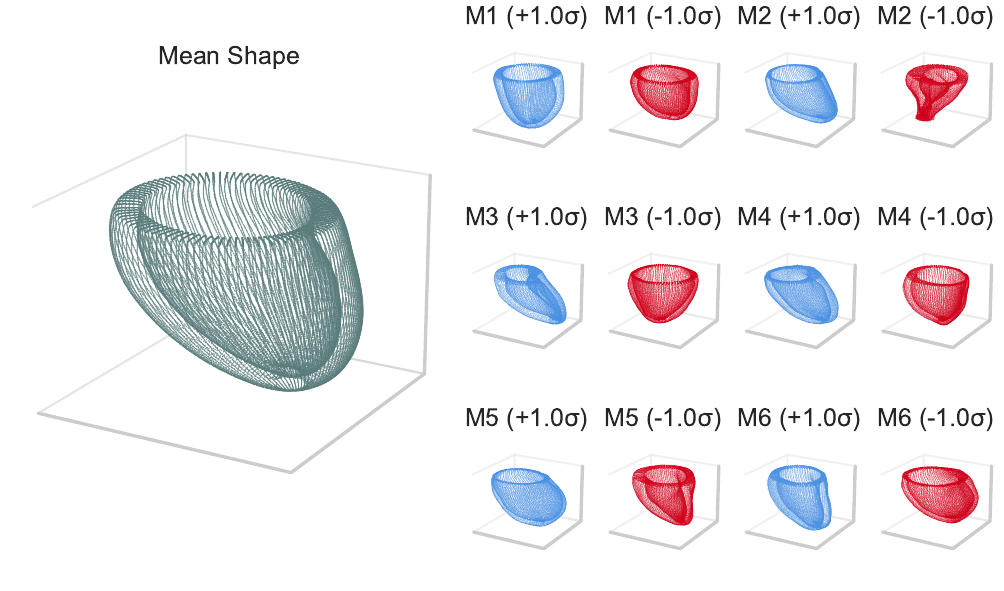}
\caption{Mean shape and first 6 principal modes of variation obtained via PCA.
The mean anatomy is shown on the left, while the first PCA modes are
visualized on the right by displaying the shapes obtained by adding
and subtracting $\sigma$ along each mode. }\label{fig:pca_modes} 
\end{figure}

Figure~\ref{fig:pca_modes} reports the mean geometry and the deformation effect of the first 6 out of 34 PCA modes. The first modes describe mainly the length of long of the long-axis, the orientation of the ventricles and the different behaviors close to the apex.

	\end{appendices}

    \bibliographystyle{ieeetr} 
    
	\bibliography{references}

\end{document}